\documentclass[10pt,twocolumn,letterpaper]{article}

\usepackage{cvpr}              % To produce the CAMERA-READY version
\usepackage{multirow}
\usepackage{colortbl}
\usepackage{xcolor}
\usepackage{pifont}
\usepackage{subcaption}

\definecolor{lightblue}{RGB}{173, 216, 230} % Light blue color
\definecolor{darkerblue}{RGB}{0, 102, 204}
\definecolor{darkblue}{RGB}{0, 0, 139}
\definecolor{lightorange}{RGB}{255, 223, 186} % Light blue color
\definecolor{forestgreen}{RGB}{34, 139, 34}
\definecolor{firebrickred}{RGB}{178, 34, 34}
\definecolor{darkorange}{RGB}{255, 140, 0}

\newcommand{\sigours}{SigLIxP}
\newcommand{\clipours}{CLIxP}

\definecolor{cvprblue}{rgb}{0.21,0.49,0.74}
\usepackage[pagebackref,breaklinks,colorlinks,allcolors=cvprblue]{hyperref}

\def\paperID{16658} % *** Enter the Paper ID here
\def\confName{CVPR}
\def\confYear{2025}

\title{Context-Aware Multimodal Pretraining}
\author{Karsten Roth$^{1,2,*}$ \quad Zeynep Akata$^{2,3}$ \quad Dima Damen$^{4}$ \quad Ivana Balažević$^{4,\dagger}$ \quad Olivier J. Hénaff$^{4,\dagger}$ \\ 
\small{$^{1}$T\"ubingen AI Center \quad $^2$Munich Center for ML \quad $^3$Helmholtz Munich, TU Munich \quad 
 $^4$Google DeepMind}\\
\small{$^*$Work done while author was at Google DeepMind. $^\dagger$Equal senior contribution.}\\ 
}
\begin{document}
\maketitle
\begin{abstract}

Large-scale multimodal representation learning successfully optimizes for zero-shot transfer at test time. Yet the standard pretraining paradigm (contrastive learning on large amounts of image-text data) does not explicitly encourage representations to support few-shot adaptation.
In this work, we propose a simple, but carefully designed extension to multimodal pretraining which enables representations to accommodate additional context.
Using this objective, we show that vision-language models can be trained to exhibit \textbf{significantly increased} few-shot adaptation: across 21 downstream tasks, we find up to four-fold improvements in test-time sample efficiency, and average few-shot adaptation gains of over 5\%, \textbf{while retaining} zero-shot generalization performance across model scales and training durations. In particular, equipped with simple, training-free, metric-based adaptation mechanisms, our representations easily surpass more complex and expensive optimization-based schemes, vastly simplifying generalization to new domains. 
% Across 21 different visual few-shot learning tasks, we find improvements in test-time sample-efficiency of over $50\%$, and absolute few-shot adaptation gains of over $4\%$, while retaining the base vision-language transfer performance across model scales and training durations.
%
% contrastive vision-language representation learning yielding %, leveraging large amounts of image-text data in order to encourage highly general 
% Shortcoming of existing works.
%
% Yet zero-shot evaluation makes no use of test-time information, and the standard pretraining paradigm (contrastive learning on large amounts of image-text data) does not encourage representations to support few-shot adaptation. 
%
% such training paradigms do not put explicit emphasis on the ability to re-use visual representations for test-time training and adaptation; assuming that zero-shot performance at scale translates into few-shot compatibility.
% Our proposal.
% In this work, we % revisit the multimodal pretraining paradigm, 
% % re-investigate this paradigm, 
%  propose a simple extension to multimodal % sigmoid-based 
%  pretraining % , $\mathbf{ctx}$SigLIP. 
\end{abstract}    
\section{Introduction}
\label{sec:intro}

%%%% BASIC MOTIVATION.

Contrastive language-image pretraining \cite{radford2021clip,jia2021align,zhai2023siglip}---where dual vision and text encoders must align both modalities while ensuring unrelated embeddings remain dissimilar---has yielded a new class of models with remarkable zero-shot transfer capabilities~\cite{radford2021clip,udandarao2022susx,zhai2023siglip,menon2023visdesc,roth2023waffleclip}.  % , allowing them to perform unseen tasks without additional training
% has catalyzed ushered a new era of large-scale visual representation learning \cite{radford2021clip,jia2021align,zhai2023siglip}. 
% % The ability to scale up image representation learning has largely been driven by advances in weakly supervised, contrastive image-text pretraining \cite{radford2021clip,jia2021align,zhai2023siglip}. 
% In these settings, dual vision and text encoders are trained to align image-text pairs while ensuring unrelated embeddings remain dissimilar.
% %match representations between
% Vision models trained using such objectives over large-scale multimodal datasets~\cite{yfcc100m,changpinyo2021cc12m,changpinyo2021conceptual,schuhmann2021laion400mopendatasetclipfiltered,schuhmann2022laionb,zhai2022lit,chen2023pali} % have become standard for downstream task transfer; as they 
% exhibit remarkable zero-shot transfer capabilities, allowing them to perform unseen tasks without additional training~\cite{radford2021clip,udandarao2022susx,zhai2023siglip,menon2023visdesc,roth2023waffleclip}.  % effectively on
%%%%% Transition into need for few-shot adaptation.
Yet even large-scale pretraining can lack out-of-distribution generalization if downstream distributions diverge from pretraining data~\cite{feuer2022captionsupervisionenablesrobust,udandarao2022susx,santurkar2023is,udandarao2024zeroshotexponentialdatapretraining,wang2024clipadapt,han2024robustclipfinetuning}.
As a result, 
additional data at test time is often essential to adapt to
% significant efforts have been made to facilitate post-hoc adaptation of vision-language models to 
% e.g. improve performance on
more severe visual distribution shifts~\cite{koh2021wildsbenchmarkinthewilddistribution,gui2024knnclip,roth2022patchcore,zhang2024dualimageenhancedclipzeroshot}, as well as more fine-grained context~\cite{cub2002011,cars196,udandarao2022susx,roth2023waffleclip,roth2024practitionersguidecontinualmultimodal}. % using additional data available at test time. % visual

\begin{figure}[t]
    \centering
    \includegraphics[width=1\linewidth]{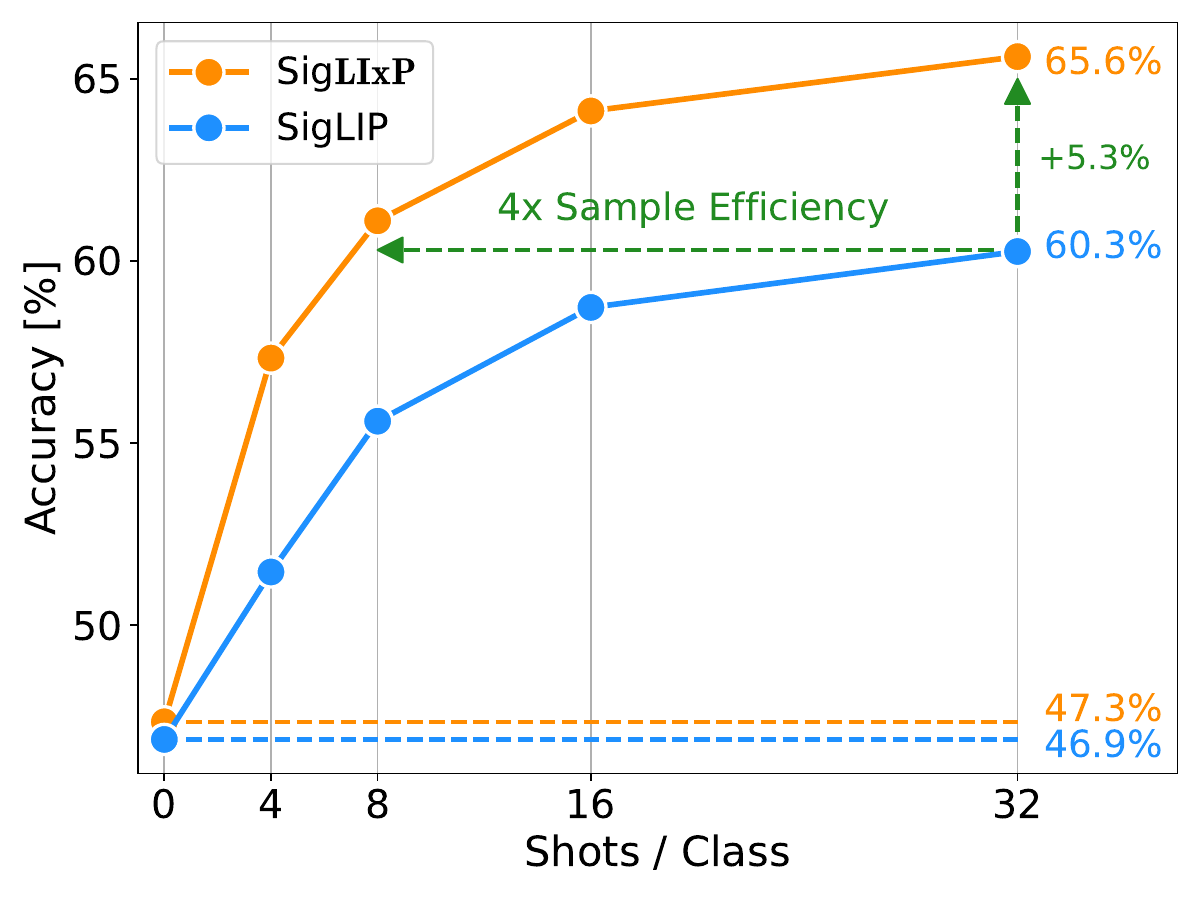}
    \vspace{-25pt}
    \caption{\textbf{Context-aware multimodal pretraining facilitates few-shot transfer.} Applying Tip-Adapter~\cite{zhang2022tipadapter} on a ViT-S/16 pretrained \textcolor{darkorange}{with} and \textcolor{darkerblue}{without} our contextualized pretraining objective (here modifying SigLIP~\cite{zhai2023siglip}) showcases \textcolor{forestgreen}{increases in test-time sample efficiency} and \textcolor{forestgreen}{overall few-shot performance} while maintaining the underlying zero-shot transfer performance.}
    \label{fig:s16_base}
    \vspace{-10pt}
\end{figure}

Such post-hoc adaptation can involve all manners of post-training optimization---including model finetuning~\cite{song2022finetune,han2024robustclipfinetuning,roth2024practitionersguidecontinualmultimodal}, prompt tuning~\cite{zhou2022cocop,thede2024reflectingstaterehearsalfreecontinual} or training of adapters~\cite{pantazis2022svladapter,wang2024clipadapt}---though generally more costly, and often prone to instability and overfitting if the number of available adaptation samples is low~\cite{fang2022data,couairon2022embeddingarithmetic,udandarao2022susx,han2024robustclipfinetuning}. At the same time, representation learning for few-shot learning on smaller models~\cite{tian2020rethinking,chen2021metabaseline,luo2023fewshot} has shown that more involved optimization-based objectives at test time and meta-training objectives during model training~\cite{lee2019metaoptnet,nichol2018reptile,xu2020optim_3} are often matched or outperformed~\cite{tian2020rethinking,chen2021metabaseline,luo2023fewshot} by simple, \textit{training-free} metric- or distance-based methods (such as prototypical or nearest-neighbor classifiers~\cite{vinyals2016metric_1,snell2017protonet,nakata2022knn,gui2024knnclip,geirhos2024flexibleperceptionvisualmemory}) operating on robust visual representations.\\

Training-free, metric-based model adaptation of large-scale vision representation models has seen widespread adoption across few- and many-shot classification, segmentation or retrieval tasks~\cite{zhang2022tipadapter,udandarao2022susx,zhou2022cocop,roth2022patchcore,balazevic2023hummingbird,gui2024knnclip,geirhos2024flexibleperceptionvisualmemory}, as it exhibits both rapid and easy scalability over both small and large numbers of  test-time examples~\cite{prabhu2023onlinecontinuallearningstorage,balazevic2023hummingbird,geirhos2024flexibleperceptionvisualmemory}, high computational efficiency e.g.\ via approximate nearest neighbor search~\cite{malkov2018ann1,johnson2021ann2}, and a great degree of flexibility in incorporating application constraints~\cite{geirhos2024flexibleperceptionvisualmemory}.
However, large-scale image-text pretraining does not explicitly account for this form of training-free model re-use at test time, instead assuming that models optimized for zero-shot generalization will inherently transfer well to few-shot scenarios~\cite{tian2020rethinking,chen2021metabaseline,luo2023fewshot,zhang2022tipadapter}.\\ 

%%%%% Our work.
In this work we question this assumption, and showcase that multimodal models can be trained to be significantly more amenable to training-free few-shot adaptation, with no reduction in their zero-shot transfer capabilities.
More precisely, this work aims to answer the question: 
\textit{``Can large-scale contrastive vision-language pretraining \textbf{better support downstream few- and many-shot transfer}, for \textbf{any metric-based adaptation} mechanism, \textbf{without impacting} zero-shot generalization performance?''}\\
%\textit{``Is it possible to modify canonical large-scale contrastive vision-language pretraining in order to \textbf{facilitate} downstream few- and many-shot transfer without making \textbf{any constraints} on the metric-based adaptation mechanism, and \textbf{without impacting} the original zero-shot generalization performance''}?

% simple, yet carefully designed
% To achieve this, we introduce the \textit{context-aware} extension to contrastive language-image pretraining, \textbf{LIxP}.

To achieve this, we introduce a simple, yet carefully designed \textit{context-aware} extension to contrastive language-image pretraining, \textbf{LIxP} (for \textbf{L}anguage-\textbf{I}mage Conte\textbf{x}tual \textbf{P}retraining). LIxP augments standard language-image contrastive objectives
%such as sigmoid~\cite{zhai2023siglip} (e.g SigLIP$\rightarrow$Sig\textbf{LIxP}) or softmax~\cite{radford2021clip} (e.g. CLIP$\rightarrow$C\textbf{LIxP}) losses
with cross-attention-based contextualization during training, preparing its representations for metric-based adaptation. Moreover, we carefully consider its inclusion into the image-text contrastive training process in order to maintain base zero-shot capabilities, leveraging particular choices in overall loss design and the use of individually learnable temperatures.
In doing so, we show that across 21 few- and many-shot downstream classification tasks LIxP objectives enable % encourage \textbf{vastly increased} few-shot adaptation efficiency 
up to \textbf{four-fold} sample-efficiency gains and over \textbf{5\%} average performance improvements, 
% \textbf{notably increased} performance ceilings 
\textit{while retaining} the original zero-shot transfer performance. % , open-vocabulary 
In doing so, LIxP closes the gap between training-free and optimization-based adaptation methods, dramatically simplifying generalization to new domains. % multimodal adaption. 
\section{Related Works}
\label{sec:related_works}

\textbf{Contrastive image-text pretraining} has become the \textit{de facto} standard when training large-scale general visual representation models. Developed initially in CLIP~\cite{radford2021clip} and ALIGN~\cite{jia2021align} using an InfoNCE-style training objective~\cite{oord2018infonce} (stemming from augmentation-based self-supervised learning such as in \cite{chen2020simclr}), several works have since augmented these in order to improve their zero-shot transfer capabilities~\cite{tejankar2021fistful,goel2021cyclip,mu2022slip,pham2023scaling}. To enable more efficient language-image pretraining at scale, SigLIP leverages only pairwise sigmoidal losses for pretraining while achieving comparable or improved transfer performance \cite{zhai2023siglip}. SigLIP and correspondingly pretrained vision-language models have seen notable adoption~\cite{evans2024badstudentsmakegreat,evans2024datacurationjointexample}. Other approaches \cite{he2019momentum,dwibedi2021nnclr,xie2023raclip,hu2023retrievalpretrain} show how external support data or training memory can be utilized to explicitly encourage increased zero-shot generalization in InfoNCE-style training. In our work, we showcase how a particular buffer and objective design choices can instead encouraging the emergence of improved few-shot capabilities while maintaining zero-shot performance.
% In this work, to conduct a large number of experiments efficiently at scale, we also employ SigLIP as our base training objective.

\noindent
\textbf{Meta- and few-shot learning} study pretraining and method design choices to facilitate adaptation to often labeled ``shots'' of new data at test time, in order to both rapidly and effectively account for changes in downstream data and label distributions. Approaches are generally separated into \textit{optimization-based} variants, which operate under the premise of full or partial model optimization over new data at test time~\cite{finn2017maml,nichol2018reptile,lee2019metaoptnet,rusu2018leo,rajeswaran2019optim_1,zintgraf2019optim_2,xu2020optim_3,triantafillou2021optim_4}, and \textit{metric-based} methods, which aim to meta-learn representation spaces equipped with a pre-defined metric~\cite{snell2017protonet,vinyals2016metric_1,chaudhuir2019metric_2,sung2018metric_3,zhang2020metric_4} to then be re-used at test time, often without any additional training. % for often training-free adaptation.
%
 % under compute constraints

While approaches are numerous, \cite{wang2019simpleshotrevisitingnearestneighborclassification,tian2020rethinking,chen2021metabaseline,li2021universal} highlight that on smaller scale, many complex meta-learning approaches are often matched or outperformed by simple metric- or regression based approaches (like prototypical or nearest neighbor classification) operating on top of a general representation backbone~\cite{luo2023fewshot}. Nevertheless, at larger scale and with increased test-time compute budgets, optimization-based approaches generally tend to produce highest performances~\cite{zhang2022tipadapter,luo2023fewshot,zhu2023ape,zhang2024dual,wu2024cascade}. 
% have shown that generalist, ImageNet pretraining does not necessarily predict few-shot classification performance in supervised scenarios; and the drawbacks in complexity when optimizing the entire model at test time.
%
% \textcolor{red}{***At large scale, optimization-based approaches are nevertheless dominant. Citations needed!!!****}
%
In this work, we ask whether well-chosen modifications to the large-scale pretraining paradigm can reduce or even close the gap between simple, cost-effective training-free methods and sophisticated optimization-based ones - while maintaining general representations capable of zero-shot transfer.  
%
% We build on these insights, and study explicit modifications in the pretraining protocol at a significantly extended scale; showing that one can actually maintain representation space generality while encouraging better training-free, few- and many-shot metric-based adaptation at test-time.\\

\noindent
\textbf{Post-hoc adaptation of large image and text representation models} such as CLIP or SigLIP has seen extensive adoption, incorporating different modifications to enhance base zero-shot transferability through architectural changes and language model extensions~\cite{zhang2024dualimageenhancedclipzeroshot,zhou2023testtime,novack2023chils,guo2023calip,pratt2023platypus,menon2023visdesc,roth2023waffleclip}, as well as test-time training through finetuning of the entire model~\cite{stojanovski2022momentumbasedweightinterpolationstrong,song2022finetune,roth2024practitionersguidecontinualmultimodal} or prompt tuning~\cite{zhou2022coop,zhou2022cocop,khattak2023promptsrc}, as well as inclusion of a learnable adapter~\cite{pantazis2022svladapter,zhang2022tipadapter,wu2024cascade}, new components~\cite{iscen2024retrievalenhanced,fifty2024contextaware,ding2024calibratedcachemodelfewshot} or generative models~\cite{kim2024datadreamfewshotguideddataset} to account for new data available at test time.
More recently, a large number of works have studied the ability to conduct such adaptation efficiently and \textit{training-free}, such as TiP-Adapter~\cite{zhang2022tipadapter}, SuS-X~\cite{udandarao2022susx} (extending TiP-Adapter with support set retrieval), Visual Memory~\cite{geirhos2024flexibleperceptionvisualmemory} utilizing large-scale visual retrieval at test time, Dual Memory~\cite{zhang2024dual} retrieving from a dual feature cache, ~\citet{zhu2023ape,ming2024fewshotretrieval} using prior refinement and retrieval augmentation against support and training data, alongside other similar extensions~\cite{wang2024clipadapt,li2024devilshotsiterativevisual,zhu2024enhancing,han2024dotadistributionaltesttimeadaptation}.
These methods rely on metric-based training-free classification mechanisms such as nearest neighbor retrieval based on the existence of a well-defined similarity metric to effectively adapt to new examples. In this work, we show how our context-aware LIxP can modify fundamental vision-language pretraining to facilitate such  general, training-free post-hoc adaptation.
% or distance-weighting 
\section{Method}
We first describe the basic contrastive language-image pretraining setup and different post-hoc, metric-based training-free adaptation mechanisms, before introducing our context-aware pretraining extension LIxP.

\subsection{Background}

\subsubsection{Contrastive Image-Text Pretraining}\label{subsubsec:contr}
For a large pretraining dataset $\mathcal{D}$ containing image-text pairs $(I_i, T_i)\!\in\!\mathcal{D}$, the softmax-training objective $\mathcal{L}_\text{CLIP}$
for a dual image  $\phi_I$~\cite{dosovitskiy2021vit} and text $\phi_T$~\cite{vaswani2017transformer} encoder is defined as
\begin{equation}\label{eq:softmax}
    \frac{1}{2|\mathcal{B}|}\sum_{i=1}^\mathcal{B}\left(\log\frac{e^{\tau_1x_it_i}}{\sum_{j=1}^{|\mathcal{B}|}e^{\tau_1x_it_h}} + \log\frac{e^{\tau_1x_it_i}}{\sum_{j=1}^{|\mathcal{B}|}e^{\tau_1x_jt_i}}\right),
\end{equation}
given a batch $\mathcal{B}$ of image-text pairs and normalized image representations $x_i = \hat{x}_i / \Vert \hat{x}_i\Vert$ with $\hat{x}_i = \phi_I(I_i)$ and textual counterparts $t_i = \phi_T(T_i) / \Vert\phi_T(T_i)\Vert$. We use $\mathbf{X}_\mathcal{B} = \phi_I(\mathcal{B}_I)$ to denote normalized image representations over the full batch of images $\mathcal{B}_I$, and $\mathbf{T}_\mathcal{B} = \phi_T(\mathcal{B}_T)$ for corresponding textual batch representations. We follow \cite{zhai2023siglip} and parameterize $\tau_1 \!=\! \exp(\tau'_1)$ with learnable  temperature $\tau'_1$. Equation~\ref{eq:softmax} uses normalization over the entire input batch $\mathcal{B}$, which significantly impacts scalability~\cite{zhai2023siglip}. Instead, \citet{zhai2023siglip} propose a pairwise sigmoid objective, which treats each image-text pair within a batch independently
\begin{equation}\label{eq:sigmoid}
    \mathcal{L}_\text{SigLIP} = -\frac{1}{|\mathcal{B}|}\sum_{i,j=1}^{|\mathcal{B}|}\log\frac{1}{1+e^{\mathbb{I}_{i=j}(-\tau_1x_it_j+b_1)}},
\end{equation}
with an indicator $\mathbb{I}_{i=j}\!=\!1$ for $i=j$, and $-$1 otherwise, and an additional learnable bias initialized at $b\!=\!-10$ \cite{zhai2023siglip}.
% For all our experiments, we leverage this SigLIP objective in \cref{eq:sigmoid} to train our vision-language models at scale. 
%In particular, we utilize the chunked implementation proposed in \cite{zhai2023siglip} for more effective multi-device training.
% Consequently, given a successfully pretrained vision-language model, these can be easily zero-shot applied to a wide range of visual, open-vocabulary classification settings: Given $N$ target classes defined with class names or descriptors $\{T_i\}_{1...N}$ and a query image $I_q$, classification is conducted by simply finding the nearest neighbor of $x_q$ in $\{t_i\}_{1...N}$.
 % to account for large optimization steps taken at the beginning of training due to the heavy imbalance towards the many negative terms (i.e. $i\neq j$)

% Training-free 
% /Many-
\subsubsection{Metric-based Few-Shot Image Classification}\label{subsec:posthoc_methods}
We consider a few- or many-shot classification task with $N$ target classes and a \textit{support set} $\mathcal{I}_\text{spt}$ of $K$ image examples per class. Setting the normalized embedded image support set as $\mathbf{X}_\text{spt}\!\in\!\mathbb{R}^{N\cdot K\times d}$ (with embedding dimensionality $d$), their corresponding labels $\mathbf{L}_\text{spt}$ and a similarity metric $s(\cdot, \cdot)$, one can define different metric-based approaches for few-shot image classification. As contrastive image-text pretraining generally operates on normalized representations, we simply set $s(\cdot, \cdot)$ to the cosine similarity. % We study six different metric-based classification methods. 
% Given L2-normalized image test features $x_\text{test} = \phi_I(I_\text{test})$, these are given as follows:

\vspace{0.5em} \noindent \textbf{Prototypical classifier~\cite{snell2017protonet}.} 
% Given the image feature support set, and the associated labels $L_\text{spt}$, 
We define a prototype representation for each class $c\in\{1,...,N\}$ as
\begin{equation}\label{eq:proto}
    \mu_c = \frac{1}{|I^c_\text{spt}|}\sum_{i\in I^c_\text{spt}}\mathbf{X}_{\text{spt},i},
\end{equation}
where $I^c_\text{spt}$ \textit{indexes} the images in the support set belonging to class $c$. Given these prototypes for each class $\mathbf{C} = [\mu_1, ..., \mu_N]$, the classification logits for the test image representation $x_\text{test}$ are computed as $x_\text{test} \mathbf{C}^T$.

\vspace{0.5em} \noindent \textbf{Tip-Adapter.}
Given text representations for all classes ${\mathbf{T}\!\in\!\mathbb{R}^{N\times d}}$,
% , the support set image features $\mathbf{S}_\text{spt}$ alongside their corresponding one-hot labels $\mathbf{L}^\text{1h}_\text{spt}$,
the Tip-Adapter logits are computed via
\begin{equation}\label{eq:tipadapter}
    \text{logits} = x_\text{test}\mathbf{T}^T + \alpha \exp\left(-\beta(1-x_\text{test}\mathbf{X}_\text{spt}^T)\right)\mathbf{L}_\text{spt}
\end{equation}
with weighting and modulation hyperparameters $\alpha$ and $\beta$. While the first term computes the standard zero-shot classification logits over textual representations, the second term reweights those against support-set image features. By default we use the original Tip-Adapter \cite{zhang2022tipadapter} hyper-parameter values $\alpha=1.0$ and $\beta=5.5$, however we also experiment with 
% \paragraph{Cross-Validated Tip-Adapter.} Initial experiments with the Tip-Adapter show that the final performance is strongly dependent on the distribution-shift between pretraining and downstream data; consequently favoring different hyperparameter combinations $\alpha$ and $\beta$. To account for this, wherever possible, we conduct
a $3-$fold cross-validation on $\mathbf{X}_\text{spt}$ to select the most favorable $\alpha, \beta$ combination, which we refer to as Cross-Validated Tip-Adapter or CV-Tip.

\vspace{0.5em} \noindent \textbf{Nearest-neighbor voting classifier.} Given the full support set $\mathbf{X}_\text{spt}$ and a test image representation $x_\text{test}$, each support example votes for its class $c_i$ with a weight $w_i$ which is proportionate to its similarity with $x_\text{test}$. Plurality-, softmax-, and rank-based voting \cite{nakata2022knn, caron2021dino,geirhos2024flexibleperceptionvisualmemory} each have their own way of deriving the weights $w_i$, see \cref{sec:supp_nn_voting}. Given the vector of weights $w$, the logits are then computed as $w \mathbf{L}_\text{spt}$.

\subsection{Context-Aware Vision-Language Pretraining}\label{subsec:contextualized_pretrain}
All methods listed in the previous section % in one way or another 
rely on metric-based aggregation and voting over a support set of normalized representations. However such re-use of supports sets for downstream metric-based classification is not explicitly accounted for in the standard contrastive pretraining setup.

\vspace{0.5em} \noindent \textbf{Representation contextualization.} To tackle this, we introduce key and value context buffers $\mathcal{M}_K$ and $\mathcal{M}_V$, which provide \textit{a proxy for test-time context during 
pretraining}. 
% allow us to convert metric-based
% to an effective surrogate training objective.
For a normalized training image representation $x_i$ as defined in \cref{subsubsec:contr}
we define its contextualized counterpart $x^\text{ctx}_i$ by cross-attending over the contextualization buffer %key and value cross-attention
\begin{equation}\label{eq:context}
    x^\text{ctx}_i = \sigma\left(\frac{x_i \cdot \mathcal{M}_K^T}{\tau_\text{ctx}\sqrt{d}}\right)\mathcal{M}_V,
\end{equation}
where $\sigma$ denotes the softmax operation across query-key similarities. We use $\mathbf{X}^\text{ctx}_\mathcal{B}$ to denote the contextualized batch representations. In all our experiments, we find it important to define $\tau_\text{ctx} \!=\! \exp(\tau'_\text{ctx})$ with learnable temperature $\tau'_\text{ctx}$.  % Crucially, f 

% Finally, contextualization can be conducted as a sequence of iterative contextualization steps $m$ with varying contextualization buffers $\mathcal{M}^m_{K,V}$, i.e.
% \begin{equation}\label{eq:successive_context}
%     x^\text{ctx}_{i,m} = \psi_\text{ctx}\left(f_\text{ctx}(x^\text{ctx}_{i,m-1}, \mathcal{M}^{m-1}_K, \mathcal{M}^{m-1}_V\right)
% \end{equation}
% where $f_\text{ctx}$ references the base contextualization in \cref{eq:context}, and $\psi_\text{ctx}$ a learnable map over contextualized representations. Experiments in \cref{subsec:ablations} reveal that a single stage of contextualization closest to the actual re-use scenario at test-time consistently works best.

\vspace{0.5em} \noindent \textbf{Contextualization while preserving zero-shot generalization.} While $\mathbf{X}^\text{ctx}_\mathcal{B}$ can be directly plugged into the contrastive loss, we find downstream few-shot adaptation gains to be limited and zero-shot transfer capabilities reduced. This can be attributed to the external memory minimizing the need to learn robustly transferable image-text representations. Thus, we explicitly separate context re-use from the representation learning objective, yielding our context-aware pretraining objective with weighting $\alpha$:
\begin{equation}
\label{eq:ctxsiglip}
    \mathcal{L}_{\textbf{LIxP}} = \alpha\mathcal{L}_\text{LIP}(\mathbf{X}_\mathcal{B},\mathbf{T}_\mathcal{B},\tau_1) + (1-\alpha)\mathcal{L}_\text{LIP}(\mathbf{X}_\mathcal{B}^\text{ctx},\mathbf{T}_\mathcal{B},\tau_2). % t_i, 
\end{equation}
Using \cref{eq:ctxsiglip}, we derive contextualized SigLIP and CLIP as \sigours \ setting $\mathcal{L}_\text{LIP} = \mathcal{L}_\text{SigLIP}$, and \clipours \ via $\mathcal{L}_\text{LIP} = \mathcal{L}_\text{CLIP}$.
Importantly, we introduce another, separately learnable temperature $\tau_2 = \exp(\tau'_2)$ to uncouple the two training objectives, which we show to be essential for optimizing for both zero- and few-shot transfer. Together, this gives three distinctly trainable temperatures: $\tau_1$, $\tau_2$ and $\tau_\text{ctx}$ (Eq~\ref{eq:final_ctx}).
% This setup is equivalently translated into our CLIP-variant, C\textbf{LIxP}:\begin{equation}
% \label{eq:ctxclip}
%     \mathcal{L}_{\text{C}\textbf{LIxP}} = \alpha\mathcal{L}_\text{CLIP}(x_i, \tau_1) + (1-\alpha)\mathcal{L}_\text{CLIP}(x^\text{ctx}_i, \tau_2). % t_i, 
% \end{equation}    
% Moreover, notice the use of a 
% which we found essential to optimize for both zero- and few-shot transfer conjointly.

\vspace{0.5em} \noindent \textbf{Contextualization buffer design.} Given our contextual mechanism and objective, the quality of the final learned representation is driven by the exact composition of % choice and integration of 
the contextualization buffers $\mathcal{M}_K$ and $\mathcal{M}_V$.
While the product between base image representation $x_i$ and $\mathcal{M}_K$ mimics the nearest neighbor retrieval process at test time, $\mathcal{M}_V$ determines the gains from context usage during training. These can be set as e.g. image or text representations of the current batch, as well as % ``stale''
% memories of 
representations from previous iterations. 
% Importantly, m
% Meaningful balance has to be struck here, since choosing e.g. $\mathcal{M}_V=\phi_T(\mathcal{B}_T)$ can introduce  shortcuts that trivially solve the pretraining loss, and significantly degrade final performance. 
% Instead 
We opt for a simple image-only context, consistent with the usage of these representations at test time:
\begin{equation}
\mathcal{M}_K = \phi_I(\mathcal{B}_I) / \Vert\phi_I(\mathcal{B}_I)\Vert,\quad \mathcal{M}_V = \phi_I(\mathcal{B}_I)
\end{equation}
% To be precise, each contextualization buffer at training step $n$ is the output of a following function:
% \begin{equation}\label{eq:buffer_design}
%     \mathcal{M}^n = \texttt{Combine}\left(f_\text{SG?}(\phi_{I,T}(\mathcal{B}_{I,T})), f_\text{prepare}(\mathcal{M}^{n-1}), \psi_\text{remap}\right)
% \end{equation}
% Here, $\phi_{I,T}$ and $\mathcal{B}_{I,T}$ denote any arbitrary combination and assignment of image and text features. $\texttt{Combine}(\cdot, \cdot, \psi_\text{remap})$ denotes the optional concatenation of current batch representations with stale representation memory from previous iterations, alongside optional remapping of each buffer entry via $\psi_\text{remap}$. Moreover, $f_\text{SG?}$ determines the option to remove $\mathcal{M}$ from the computational graph, while $f_\text{prepare}$ covers various forms of buffer preparation such as LayerNormalization and remapping using a special buffer map, as e.g. done in Hummingbird~\cite{balazevic2023hummingbird}.
% our work, particularly 
While $\mathcal{M}_K$ contains normalized entries as expected for retrieval at test-time, $\mathcal{M}_V$ utilizes \textit{non-normalized} embeddings to leverage the additional degree of freedom in the norm \cite{scott2011vmf,kirchhof2022nonisotropic} to provide an additional value signal leverageable during training.
In \cref{subsec:ablations}, we thoroughly evaluate alternative choices for buffer design, and show that this simple scalable setup outperforms alternatives. Moreover, having equivalence between the buffer and the training batch notably improves compute efficiency.
% defined simply as $\mathcal{M}_K = \mathcal{M}_V = \phi_I(\mathcal{B}_I)$. 
Our extensive experiments across model sizes reveal that for metric-based downstream adaptation, allowing the model to jointly populate and backpropagate through the buffer with image representations is key to encourage re-use at test time. Deviations from this basic setup consistently fail to strike a suitable tradeoff between maintaining zero-shot generalization performance and improving few-shot adaptation performance.

\vspace{0.5em} \noindent \textbf{Overall objective.} Put together, our training objective is defined as the single-stage application of \cref{eq:context} with $\mathcal{M}_K = \mathbf{X}_\mathcal{B} = \phi_I(\mathcal{B}_I)/\Vert\phi_I(\mathcal{B}_I)\Vert$ and $\mathcal{M}_V = \hat{\mathbf{X}}_\mathcal{B} = \phi_I(\mathcal{B}_I)$, s.t. for batch $\mathcal{B}_I$ we have
\begin{equation}\label{eq:final_ctx}
    \mathbf{X}^\text{ctx}_\mathcal{B} = \frac{\hat{\mathbf{X}}^\text{ctx}_\mathcal{B}}{\Vert \hat{\mathbf{X}}^\text{ctx}_\mathcal{B} \Vert}\text{,} \quad \hat{\mathbf{X}}^\text{ctx}_\mathcal{B} = \sigma\left(\frac{\mathbf{M} \odot \mathbf{X}_\mathcal{B}\mathbf{X}_\mathcal{B}^T}{\tau_\text{ctx}\sqrt{d}}\right)\hat{\mathbf{X}}_\mathcal{B}
\end{equation}
where $\odot$ denotes element-wise multiplication. Importantly, $\mathbf{M} = \mathbf{1} - \mathbf{I}_{\infty}$, with $\mathbf{I}_{\infty}$ denoting diagonal ``$\infty$'' entries, defines a ones-mask with $-\infty$-diagonals to avoid a representation attending to itself (i.e. setting softmax entries for self-attention to $0$) and reverting to %in order to enforce context usage; instead of falling into the shortcut of simple attending to itself and solving 
the standard image-text training objective.
For a large enough batch-size, this effectively creates implicit per-iteration episodic training; where each non-masked batch entry constitutes support samples used to predict respective textual embeddings. % for both SigLIP objectives, as well as final contextualization operation.

\section{Experiments}

Our pretraining pipeline follows the SigLIP~\cite{zhai2023siglip} protocols for training ViT-(S/16, B/16, L/16) image encoders and correspondingly sized BERT-\{S,B,L\} text encoders. To evaluate both zero-shot as well as few-shot transfer capabilities, we measure performance on 21 diverse datasets commonly used for few-shot and domain adaptation (see \cref{sec:supp_experimental_details}). %: CUB200-2011~\cite{cub2002011}, Stanford Cars~\cite{cars196}, Cassave~\cite{cassava}, CIFAR100~\cite{cifar100}, Colorectal Histology~\cite{colhist}, DomainNet-$\{$ClipArt, Infograph, Quickdraw, Sketch$\}$~\cite{domainnet}, DTD~\cite{dtd}, EuroSAT~\cite{eurosat}, Food101~\cite{food101}, ImageNet2012~\cite{imagenet2012}, ImageNet-Sketch~\cite{imagenetsketch}, Oxford IIIT Pets~\cite{pets}, Places365~\cite{places365}, Plant-Village~\cite{plantvillage}, RESISC45~\cite{resisc45}, Stanford Dogs~\cite{dogs}, SUN397~\cite{sun397} and UC Merced~\cite{ucmerced}.
% For datasets where train/test or train/validation splits are not available, we introduce suitable splits with adaptation examples (more information in the supplementary).
% leverages the \texttt{big\_vision} codebase~\cite{bigvision}, and
% We train different variations of vision transformer image encoders
The majority of our experiments use the SigLIP pretraining objective due to its superior computational efficiency, but we show all our conclusions hold for CLIP in \cref{tab:clip_arch_comp}.

\subsection{Context-Aware Training}

\begin{figure}[t!]
    \centering
    \includegraphics[width=1\linewidth]{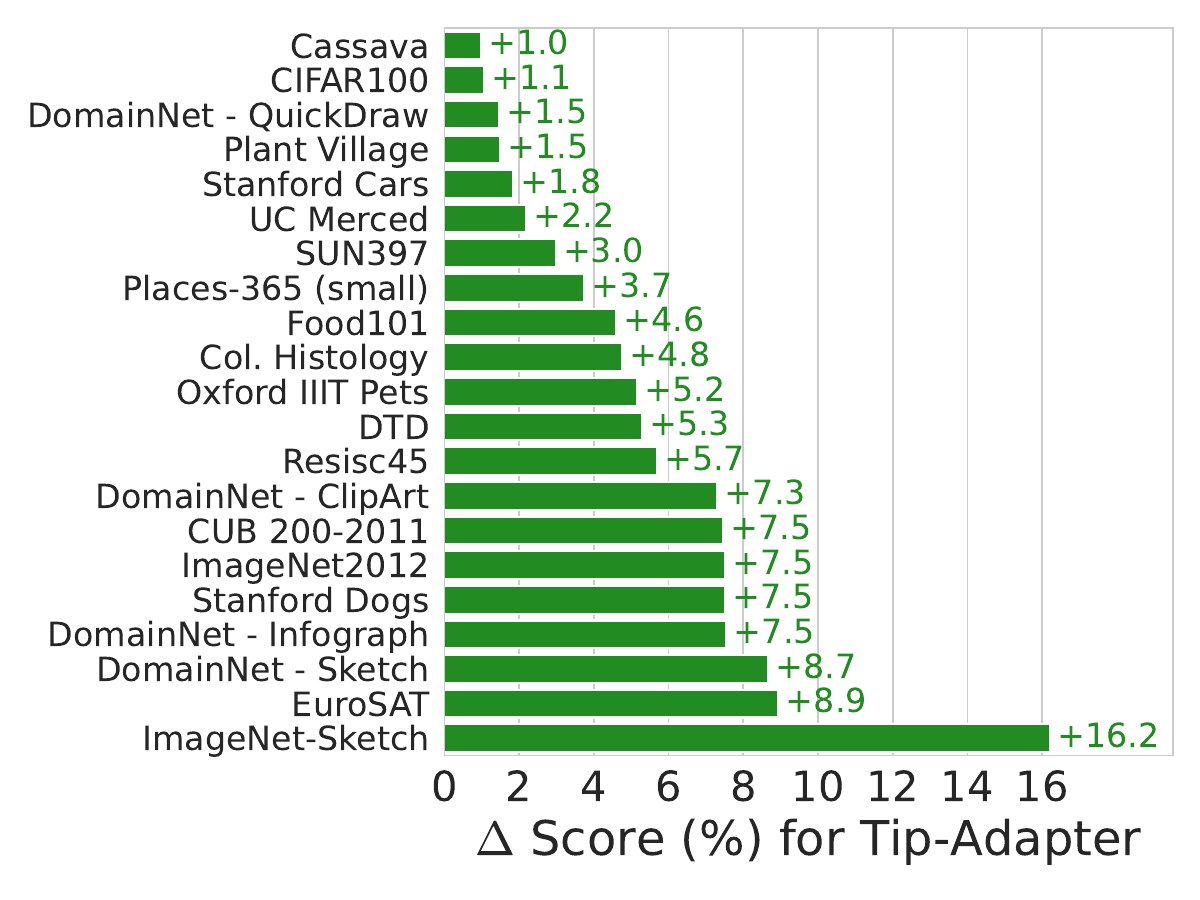}
    \vspace{-25pt}
    \caption{\textbf{Dataset-level \textcolor{darkorange}{performance breakdown} (32-shot, Tip-Adapter~\cite{zhang2022tipadapter})} for ViT-S/16 shows gains up to \textcolor{forestgreen}{$+16.2\%$} on \textbf{all 21} benchmarks; with each dataset improving by at least \textcolor{forestgreen}{$+1.0\%$}.}
    \label{fig:dataset_breakdown}
    \vspace{-10pt}
\end{figure}

\vspace{0.5em} \noindent \textbf{Impact on downstream few-shot adaptation.} To study the impact of context-aware pretraining on downstream few-shot adaptation, we apply \sigours \ pretraining to a ViT-S/16 model (57.2M parameters) trained over 1.5B WebLI examples and evaluate its few-shot adaptation performance with the Tip-Adapter (see \cref{eq:tipadapter}). Results are shown in \cref{fig:s16_base}, where we compare zero- to 32-shot adaptation performance for \textcolor{darkorange}{\sigours}\ against an equivalently pretrained \textcolor{darkerblue}{SigLIP} model.
% and the only dataset with no notable improvement being \texttt{Places365}.
%
% (same model, seed and training example count). 
% Dashed lines denote zero-shot transfer performance averaged over all datasets in both cases. 
%
While zero-shot performance remains comparable (\textcolor{forestgreen}{$+0.4\%$} improvements), we find significant gains in few-shot performance (e.g. \textcolor{forestgreen}{$+5.3\%$} for 32-shots) and a 4$\times$ increase in sample efficiency (\sigours \ achieving \textcolor{darkorange}{$61.1\%$ 8-shot} vs SigLIP \textcolor{darkerblue}{$60.3\%$ 32-shot}). Looking at a dataset-level breakdown in \cref{fig:dataset_breakdown} we find that all 21 datasets exhibit over \textcolor{forestgreen}{+1\%} gains, with a maximum gain of \textcolor{forestgreen}{+16.2\%} on ImageNet-Sketch.
% This improvements are highly noteworthy, and already on a smaller scale allow for vision-backbones to be trained for much improved few-shot performance with no detriment to zero-shot transfer performance.
\begin{figure}[t!]
    \centering
    \includegraphics[width=1\linewidth]{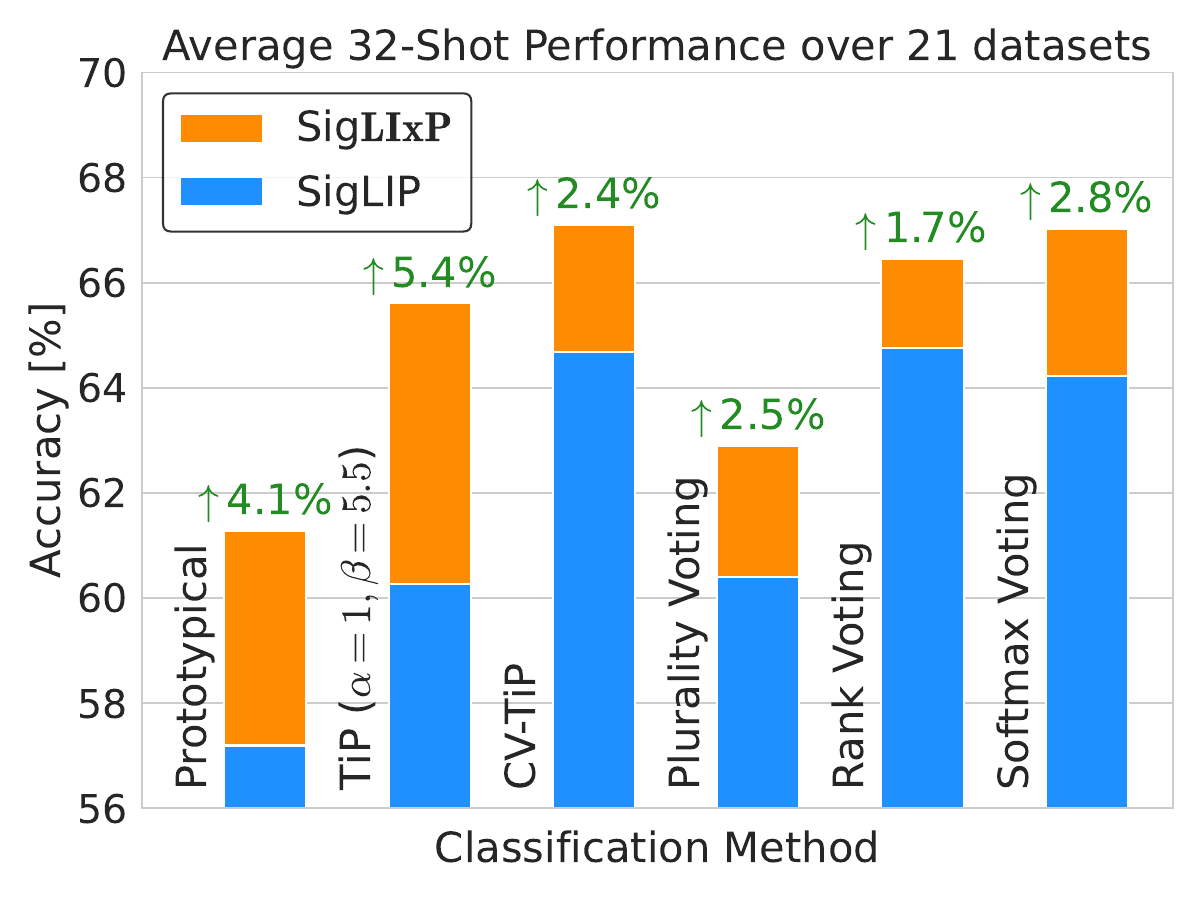}
    \vspace{-20pt}
    \caption{\textbf{Significant gains across metric-based few-shot classifiers.} Applying prototypical classification~\cite{snell2017protonet}, Tip-Adapters~\cite{zhang2022tipadapter} and nearest neighbor classifiers~\cite{nakata2022knn,geirhos2024flexibleperceptionvisualmemory} on vision-backbones using \textcolor{darkorange}{context-aware pretraining} significantly boosts $32$-shot results across the board (here ViT-S/16, 1.5B ex.).}
    \label{fig:posthoc_comparison}
    \vspace{-8pt}
\end{figure}
% \noindent
To additionaly showcase the \textit{robustness of} LIxP \textit{to the choice of the few-shot adaptation method}, we consider all six metric-based adaptation methods described in \cref{subsec:posthoc_methods} (see \cref{fig:posthoc_comparison}). In all cases, improvements remain high, ranging from a \textcolor{forestgreen}{$+1.7\%$} gain for \textit{distance-free} rank-voting to \textcolor{forestgreen}{$+5.4\%$} for Tip-Adapter. These results indicate that \textbf{LIxP} offers a general solution to facilitate training-free adaptation at test time, allowing practitioners to choose adaptation methods based on individual constraints.
% configuration ($\alpha=1.0, \beta=5.5$). 
%Overall highest \textcolor{darkerblue}{base performance} is achieved using cross-validated Tip-Adapter, Rank and Softmax Voting; despite \textcolor{darkorange}{contextualized pretraining} gains being highest for the cross-validated Tip-Adapter and Softmax voting.
% ; both of which explicitly incorporate distance weighting. 
% Finally, we note that the notable gains on Tip-Adapter variations (which operate on a combination of zero-shot and few-shot classification) further highlight that \textcolor{darkorange}{$\mathbf{ctx}$SigLIP} manages to maintain zero-shot transferability while consistently improving on few-shot re-use. 
% Going forward, if not specified otherwise, we report results on the \textit{prototypical}, \textit{cross-validated Tip-Adapter} and \textit{softmax-voted nearest neighbor} classification for post-hoc adaptation method diversity and to cover the highest performing method.

\label{subsubsec:scale}
\begin{table}[t!]
    \centering
    \resizebox{1\linewidth}{!}{
    \begin{tabular}{l|ccccc}
        \textbf{Model} $\rightarrow$ & ViT-S/16 & $\rightarrow$ & ViT-B/16 & $\rightarrow$ & ViT-L/16\\
        \textbf{Examples} $\rightarrow$ & 1.5B & 6B & 6B & 15B & 8B\\
        \midrule
        \multirow{2}{*}{ZeroShot} & $46.9$ & $52.1$ & $60.3$ & $62.5$ & $64.1$\\
        & \textcolor{forestgreen}{$+0.4$} & \textcolor{firebrickred}{$-0.2$} & \textcolor{firebrickred}{$-0.4$} & \textcolor{firebrickred}{$-0.5$} & \textcolor{firebrickred}{$-0.1$}\\
        \multirow{2}{*}{Prototypical} & $57.2 \pm 0.2$ & $60.8 \pm 0.3$ & $66.8 \pm 0.2$ & $67.4 \pm 0.3$ & $70.7 \pm 0.3$\\
        & \textcolor{forestgreen}{$+4.1$} & \textcolor{forestgreen}{$+3.4$} & \textcolor{forestgreen}{$+3.4$} & \textcolor{forestgreen}{$+3.9$}& \textcolor{forestgreen}{$+3.3$}\\
        %%%%%%
        \multirow{2}{*}{Default Tip} & $60.3 \pm 0.1$ & $63.6 \pm 0.3$ & $69.5 \pm 0.2$ & $70.2 \pm 0.3$ & $73.2 \pm 0.3$\\
        & \textcolor{forestgreen}{$+5.4$} & \textcolor{forestgreen}{$+4.8$} & \textcolor{forestgreen}{$+4.3$} & \textcolor{forestgreen}{$+4.3$}& \textcolor{forestgreen}{$+4.0$}\\
        %%%%%%
        \multirow{2}{*}{XVal Tip} & $64.7 \pm 0.2$ & $67.8 \pm 0.3$ & $73.8 \pm 0.2$ & $74.7 \pm 0.2$ & $77.0 \pm 0.3$\\
        & \textcolor{forestgreen}{$+2.4$} & \textcolor{forestgreen}{$+2.3$} & \textcolor{forestgreen}{$+1.6$} & \textcolor{forestgreen}{$+1.6$} & \textcolor{forestgreen}{$+1.4$}\\
        %%%%%%
        \multirow{2}{*}{Plurality NN} & $60.4 \pm 0.1$ & $63.8 \pm 0.2$ & $69.1 \pm 0.2$ & $69.6 \pm 0.3$ & $72.6 \pm 0.2$\\
        & \textcolor{forestgreen}{$+2.5$} & \textcolor{forestgreen}{$+1.9$} & \textcolor{forestgreen}{$+2.3$} & \textcolor{forestgreen}{$+2.6$} & \textcolor{forestgreen}{$+1.8$}\\
        %%%%%%
        \multirow{2}{*}{Rank NN} & $64.8 \pm 0.1$ & $68.1 \pm 0.1$ & $73.2 \pm 0.1$ & $74.1 \pm 0.2$ & $76.5 \pm 0.2$\\
        & \textcolor{forestgreen}{$+1.7$} & \textcolor{forestgreen}{$+1.3$} & \textcolor{forestgreen}{$+1.8$} & \textcolor{forestgreen}{$+1.8$} & \textcolor{forestgreen}{$+1.1$}\\
        %%%%%%
        \multirow{2}{*}{Softmax NN} & $64.2 \pm 0.1$ & $67.5 \pm 0.2$ & $72.6 \pm 0.2$ & $73.3 \pm 0.3$ & $75.9 \pm 0.2$\\
        & \textcolor{forestgreen}{$+2.8$} & \textcolor{forestgreen}{$+2.4$} & \textcolor{forestgreen}{$+2.6$} & \textcolor{forestgreen}{$+2.7$} & \textcolor{forestgreen}{$+1.8$}\\
        %%%%%%
        \midrule
        \textbf{Average Gain} & \textcolor{forestgreen}{$\mathbf{+3.2}$} & \textcolor{forestgreen}{$\mathbf{+2.7}$} & \textcolor{forestgreen}{$\mathbf{+2.7}$} & \textcolor{forestgreen}{$\mathbf{+2.8}$} & \textcolor{forestgreen}{$\mathbf{+2.2}$}\\
    \end{tabular}}
    \caption{\textbf{Context-aware pretraining for SigLIP holds across model size and data scale.} Benefits of contextualized pretraining across larger architectures, longer pretraining runs, and combinations of both (here shown for $32$-shot classification for three metric-based classifiers) show that our contextualized pretraining objective boosts performance across model sizes and training duration; with no or at most very marginal drops in zero-shot transfer.}
    \label{tab:arch_comp}
    \vspace{-15pt}
\end{table}

\vspace{0.5em} \noindent \textbf{Scaling model and data size.}
To ensure these insights hold across increased model and training data sizes, we further consider ViT-S/16 trained on 6B examples, ViT-B/16 on 6B and 15B examples, and ViT-L/16 on 8B examples in \cref{tab:arch_comp}.
% extends both training data, model size as well both both in conjunction; going to . 
% All approaches utilize the same training paradigm, contrasting gains against a directly comparable SigLIP run.
Improvements remain high across the board, showcasing the \textit{robustness of our approach to the size of the model and the amount of training data used}.  
For example, training ViT-S/16 for $4\times$ the amount of examples improves base zero-shot performance by $5.2\%$, while gains from switching to \textcolor{darkorange}{context-aware pretraining} remain high and stable (e.g.~\textcolor{forestgreen}{$+2.4\%$}$\rightarrow$\textcolor{forestgreen}{$2.3\%$} for the cross-validated Tip-Adapter). Similarly, training a ViT-B/16 for 15B versus 6B training examples again maintains relative gains (e.g. \textcolor{forestgreen}{$+2.6\%$}$\rightarrow$\textcolor{forestgreen}{$2.7\%$} when used with softmax-voted nearest neighbor classification). 
% These gains also remain fairly comparable to these exhibited by the smaller scale ViT-S/16, where for both 6B example pretrained variations, gains using softmax-voted nearest neighbor classification are \textcolor{forestgreen}{$+2.4\%$} and \textcolor{forestgreen}{$+2.6\%$}, respectively. 
When scaling model and data further to ViT-L/16 (8B), improvements remain high (e.g. \textcolor{forestgreen}{$+1.8\%$} for SNN and \textcolor{forestgreen}{$+4.0\%$} for Tip-Adapter), with slightly lower absolute gains in parts attributable to several datasets reaching classification performance plateaus (e.g. $95.2\%$ on \texttt{UC Merced}): When excluding datasets with \textcolor{darkerblue}{base 32-shot performances} over e.g $70\%$, the average gain for example on Tip-Adapter improves from \textcolor{forestgreen}{$+4.0\%$} to \textcolor{forestgreen}{$+5.9\%$}; with the same performance thresholding on the smallest ViT-S/16 (1.5B) giving only a \textcolor{forestgreen}{$+5.4\%$}$\rightarrow$\textcolor{forestgreen}{$+5.8\%$} jump.
% , but then e.g. \texttt{Stanford Cars} $92.4\%\rightarrow94.3\%$, \texttt{Food101} $88.2\%\rightarrow92.7\%$ or \texttt{Oxford IIIT Pets} $86.2\%\rightarrow91.5\%$).

\begin{figure}[t!]
    \centering
    \includegraphics[width=1\linewidth]{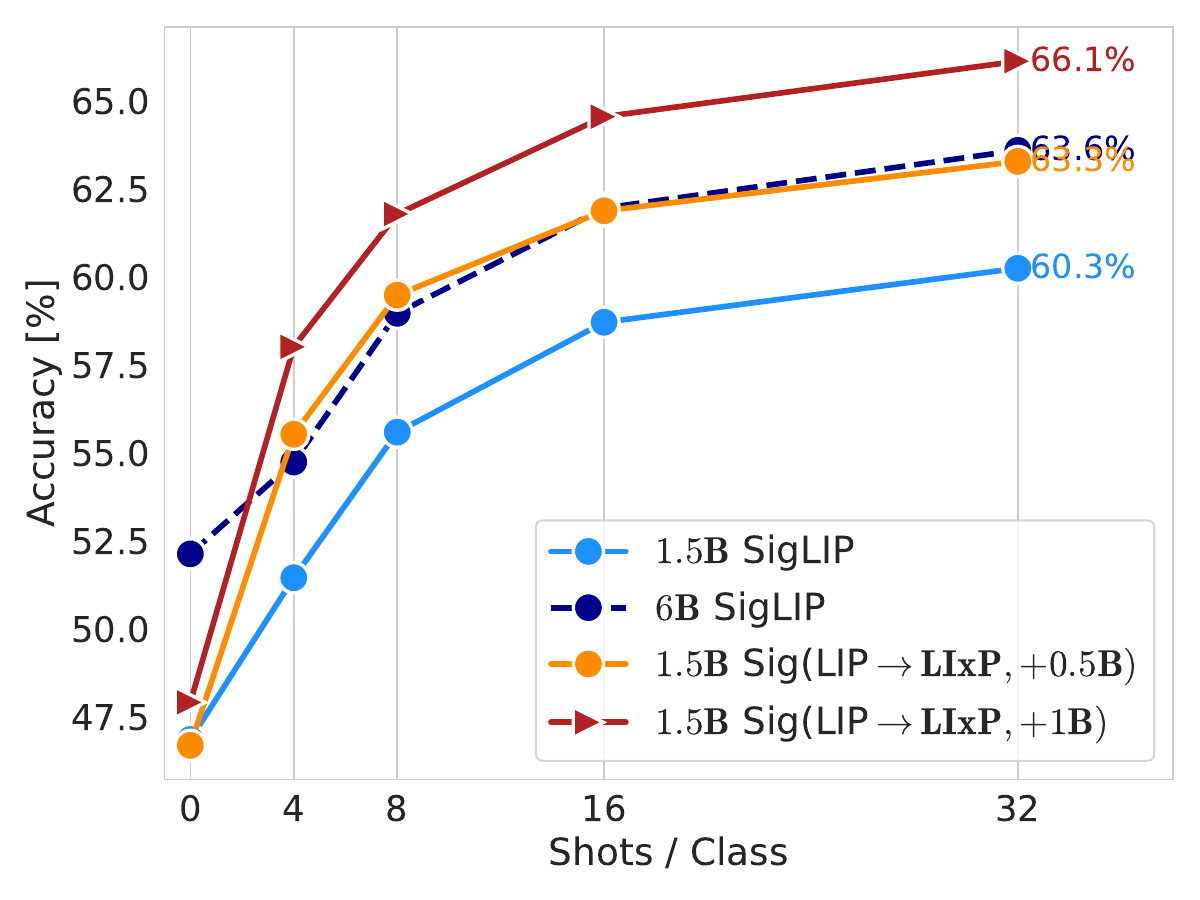}
    \vspace{-20pt}
    \caption{\textbf{Context-aware post-training.} We apply Sig\textbf{LIxP} on an already \textcolor{darkerblue}{pretrained ViT-S/16 (1.5B examples)}, finetuning for \textcolor{darkorange}{+0.5B} and \textcolor{firebrickred}{+1B} examples. We contrast the performance against a \textcolor{darkblue}{6B ViT-S/16}. Results indicate that context-aware finetuning can match much \textcolor{darkblue}{longer base pretraining} with only \textcolor{darkorange}{+0.5B} examples, and noticeably outperform it with just \textcolor{firebrickred}{+1B} examples - even if the base zero-shot transfer performance of the \textcolor{darkblue}{6B reference model} is much higher. Visualized results use Tip-Adapter for classification.}
    \label{fig:post_training}
\end{figure}

\vspace{0.5em} \noindent \textbf{Context-aware post-training.} Finally, we study the possibility of contextualized post-training, where \sigours \ is applied after base SigLIP pretraining. Results on a ViT-S/16 model pretrained on 1.5B WebLI examples are shown in \cref{fig:post_training}. When finetuning on \textcolor{darkorange}{+0.5B} additional examples using \sigours, zero-shot performance is retained but 
%an improvement in the few-shot adaptation performance --- 
 $32$-shot % \textcolor{darkerblue}{baseline } 
performance increases from \textcolor{darkerblue}{60.3\%} to \textcolor{darkorange}{63.3\%}, even approaching the \textcolor{darkblue}{6B example baseline} (trained on $3\times$ more examples).
% , which achieves \textcolor{darkblue}{$60.8\%\pm0.3$}. 
% (c.f. also \cref{tab:arch_comp}). 
% This is particularly noteworthy considering that the \textcolor{darkblue}{6B} SigLIP few-shot performance benefits from an increase in zero-shot performance by over $5\%$ compared to \textcolor{darkblue}{6B} SigLIP. 
When finetuning on \textcolor{firebrickred}{+1B} examples, we outperform the 6B SigLIP baseline (\textcolor{darkblue}{$63.6\%$} vs \textcolor{firebrickred}{$66.1\%$}).
% , with only small gains in zero-shot performance. 
% Interestingly, post-hoc contextualized finetuning on \textcolor{firebrickred}{+1B} examples outperforms full contextualized pretraining on 1.5B examples (c.f. \cref{tab:arch_comp}); in parts attributable to the emergent nature of full contexualized pretraining described in \cref{subsubsec:exp_training}. 
These results clearly indicate that post-hoc contextualized finetuning is possible and, measured by number of contextualized training iterations, more sample-efficient. Moreover, \textcolor{firebrickred}{contextualized finetuning} can significantly improve on few-shot adaptation performance 
% with significantly fewer examples 
compared to \textcolor{darkblue}{baseline pretraining} which uses more than twice the number of examples. 
% We do find that for a model to acquire improved few-shot adaptability through contextualized finetuning, a minimum number of finetuning steps are required: Adding only 0.1B contextualized pretraining examples only raises $32$-shot adaptation performance from \textcolor{darkerblue}{$57.2\%$} to $57.9\%$.

\begin{figure}[t!]
    \centering
    \includegraphics[width=1\linewidth]{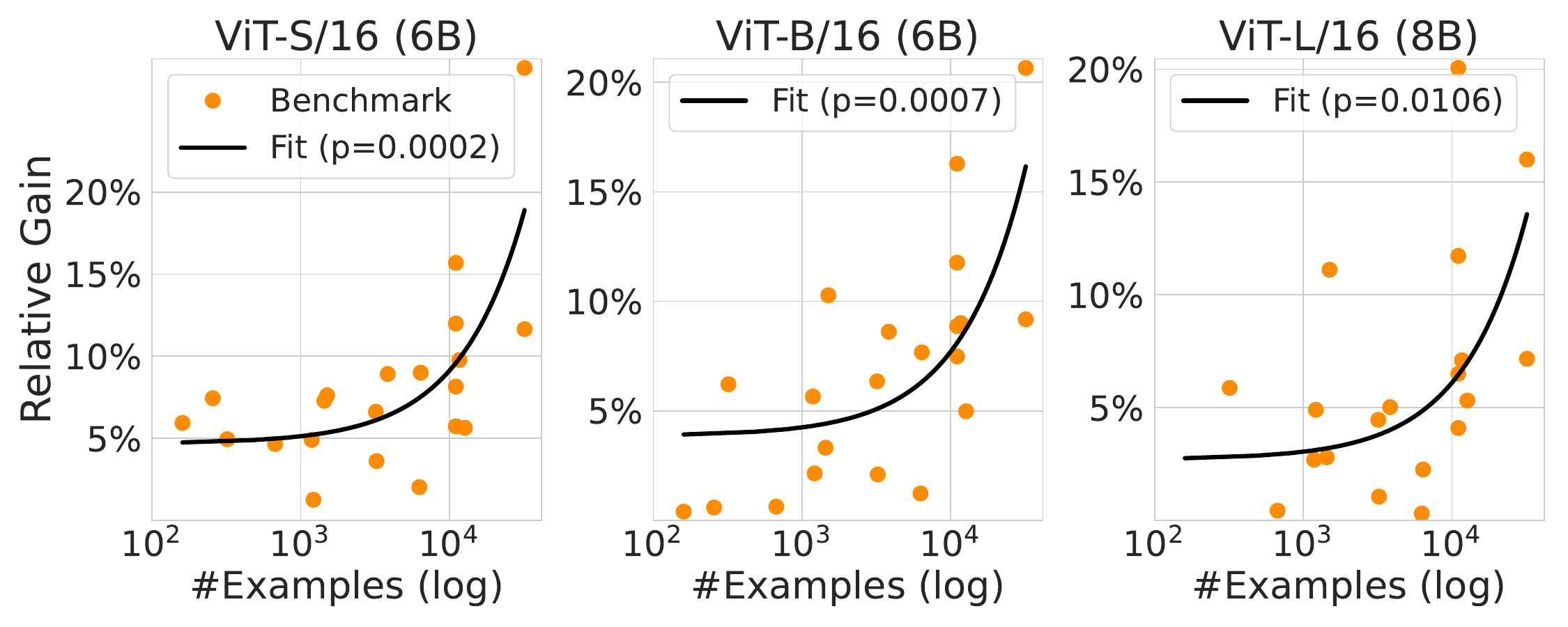}
    \vspace{-15pt}
    \caption{\textbf{Contextualized pretraining particularly benefits many-shot transfer.} For all 21 evaluation benchmarks, we plot the absolute number of examples (shots/class $\times$ $\#$classes) against the relative gain when switching to \sigours. Results shown are for 32 shots/class. We find consistent relative gain in all scenarios, which become higher as absolute example counts increases.}
    \label{fig:relative_gains}
    \vspace{-8pt}
\end{figure}

\vspace{0.5em} \noindent \textbf{Benefits across few- and many-shots.}
% Our results indicate that contextualized pretraining results in significant gains across all evaluation benchmarks. 
Since each dataset has a different number of classes $N$, the actual number of support examples $N \times K$ varies per dataset.  To account for this, we visualize relative gains (\sigours \ $-$ SigLIP) / SigLIP over baseline training as a function of absolute number of examples provided in \Cref{fig:relative_gains}. The linear fit (visualized semi-logarithmically) and p-value estimates are computed using \cite{seabold2010statsmodels}. Even when taking certain confounds (such as high base performance) into account, we find that performance gains increase  with the absolute number of examples and reach their maximum in the many-shot setup. 

\begin{table}[t!]
    \centering
    \resizebox{1\linewidth}{!}{
    \begin{tabular}{lc|ccccc}
        \textbf{Method} & \textbf{Train-free} & \textbf{IN-1K} & \textbf{DTD} & \textbf{Food101} & \textbf{Pets} & \textbf{Cars}\\
        \midrule
        Linear Probe~\cite{wu2024cascade} & \textcolor{red}{\ding{55}} & 67.3 & 70.0 & 82.9 & 85.3 & 80.4\\
        TIP-X~\cite{udandarao2022susx} & \textcolor{forestgreen}{\ding{51}} & 71.1 & - & - & - & - \\ 
        % Tip-Adapter~\cite{zhang2022tipadapter} &  & \ding{51} & \\
        % CoCoOp~\cite{zhou2022cocop} & \ding{55} & 70.8 & 63.0 & 87.3 & 71.6 & 93.3 \\
        % CoOp~\cite{zhou2022coop} & \ding{55} & 71.9 & 69.9 & 84.2 & 91.9& 83.1 \\
        APE~\cite{zhu2023ape} & \textcolor{forestgreen}{\ding{51}} & 72.1 & - & - & - & - \\
        DMN-TF~\cite{zhang2024dual} & \textcolor{forestgreen}{\ding{51}} & 72.6 & 71.9 & 86.0 & 92.9 & 78.4 \\
        Clip-Adapter~\cite{gao2021clipadapter} & \textcolor{red}{\ding{55}} & 71.1 & - & - & -\\
        MaPLe~\cite{khattak2023maple} & \textcolor{red}{\ding{55}} & 72.3 & 71.3 & 85.3 & 92.8 & 83.6\\
        PromptSRC~\cite{khattak2023promptsrc} & \textcolor{red}{\ding{55}} & 73.2 & 72.7 & 87.5 & 93.7 & 83.8\\
        Tip-Adapter-F~\cite{zhang2022tipadapter} & \textcolor{red}{\ding{55}} & 73.7 & - & - & - & - \\
        APE-T~\cite{zhu2023ape} & \textcolor{red}{\ding{55}} & 74.3 & - & - & - & -\\
        CasPL~\cite{wu2024cascade} & \textcolor{red}{\ding{55}} & 74.2 & 75.1 & 88.4 & 94.1 & 86.7 \\
        DMN~\cite{zhang2024dual} & \textcolor{red}{\ding{55}} & 74.7 & 75.0 & 87.1 & 94.1 & 85.3 \\
        \midrule
        % \textcolor{darkorange}{Sig\textbf{LIxP}}: CVTip & \ding{51} & 77.2 & 75.7 & 92.3 & 94.0 & \textbf{93.6} \\
        \textcolor{darkorange}{Sig\textbf{LIxP}} & \textcolor{forestgreen}{\ding{51}} & \textbf{77.9} & \textbf{76.7} & \textbf{92.6} & \textbf{94.4} & \textbf{92.8} \\ % : ZS+SNN
    \end{tabular}}
    \vspace{-7pt}
    \caption{\textbf{Comparison against optimization-based methods in literature.} On five standard benchmarks, we compare Sig\textbf{LIxP} with simple \textit{training-free}, softmax-voted nearest-neighbor classifier on top of the base zero-shot classification against e.g. finetuned Tip-Adapter~\cite{zhang2022tipadapter} or SOTA prompt-learning methods~\cite{khattak2023promptsrc,wu2024cascade} (16-shot) on the same backbone (ViT-B/16). We show that through context-aware pretraining, simple \textit{training-free} metric classifiers can be made highly competitive, strongly outperforming specifically tuned optimization approaches. %\vspace{-1em} % While orthogonal, w
    }
    \label{tab:comparison_learn_literature}
    \vspace{-10pt}
\end{table}
% \begin{table}[t!]
%     \centering
%     \resizebox{1\linewidth}{!}{
%     \begin{tabular}{l|ccccc}
%         \textbf{Method} & IN-1K & EuroSAT & DTD & Pets & Food101\\
%         % \midrule
%         % \multicolumn{6}{c}{ViT-B/16}\\
%         % \midrule
%         CoOp~\cite{zhou2022cocop} & 71.9 & & & & \\
%         Clip-Adapter~\cite{gao2021clipadapter} & 71.1& & & & \\
%         Tip-Adapter~\cite{zhang2022tipadapter} (B/16) & 70.8 & & & & \\
%         Tip-F~\cite{zhang2022tipadapter}$^*$ (B/16) & 73.7 & & & & \\
%         \midrule
%         \multicolumn{6}{c}{Other backbones}\\
%         \midrule        
%         Tip-X~\cite{udandarao2022susx} & & & & & \\
%          & & & & & \\
%     \end{tabular}}
%     \caption{We contrast a Sig\textbf{LIxP}-pretrained ViT-B/16 against other literature results on exact dataset splits using the setup described in \cite{zhang2022tipadapter,udandarao2022susx}.}
%     \label{tab:direction_comparison}
% \end{table}

\vspace{0.5em} \noindent \textbf{Comparison with state-of-the-art.} In \cref{tab:comparison_learn_literature}, we compare \sigours \ results to state-of-the-art finetuning and prompt-learning methods on the same ViT-B/16 model.
These methods specifically finetune and optimize for the downstream domains at hand (on 16-shots/class).
In contrast, our \sigours \ backbone equipped with a simple and highly scalable \textit{training-free} mechanism (softmax NN + zero-shot logits) strongly outperforms these state-of-the-art optimization-based adaptation schemes, 
%These results are not meant to be directly comparable, but rather aim to showcase the simplicity in 
% We outperform learning-based approaches simply through changes in the pretraining paradigm - achieving 
achieving \textcolor{darkorange}{$77.9\%$} on ImageNet 16-shot (versus state-of-the-art DMN~\cite{zhang2024dual} $74.7\%$), or \textcolor{darkorange}{$76.7\%$} on DTD versus state-of-the-art CasPL~\cite{wu2024cascade} $75.1\%$.
Our context-aware pretraining \textit{closes the gap between optimization-based and training-free methods}, where the latter generally lag behind the more sophisticated and expensive optimization-based ones, thereby notably simplifying adaptation to new domains, as a single unmodified model can be now used cheaply across domains. 
% In doing so, %These results showcase how 

% , and strongly outperform prior training-free methods. 

% as our base contrastive language-image

\begin{table}[t!]
    \centering
    \resizebox{1\linewidth}{!}{
    \begin{tabular}{l|ccccc}
        \textbf{Model} $\rightarrow$ & ViT-S/16 & $\rightarrow$ & $\rightarrow$ & ViT-B/16 & $\rightarrow$\\
        \textbf{Examples} $\rightarrow$ & 1.5B & 6B & 15B & 6B & 15B\\
        \midrule
        \multirow{2}{*}{ZeroShot} & $47.9$ & $52.1$ & $53.3$ & $60.8$ & $61.9$\\
        & \textcolor{firebrickred}{$-0.2$} & \textcolor{forestgreen}{$0.0$} & \textcolor{firebrickred}{$-0.1$} & \textcolor{firebrickred}{$-0.2$} & \textcolor{forestgreen}{$+0.5$}\\
        %%%%%%%
        \multirow{2}{*}{Prototypical} & $57.0 \pm 0.3$ & $59.5 \pm 0.3$ & $61.5 \pm 0.3$ & $66.3 \pm 0.3$ & $67.3 \pm 0.2$\\
        & \textcolor{forestgreen}{$+4.0$} & \textcolor{forestgreen}{$+5.0$} & \textcolor{forestgreen}{$+4.0$} & \textcolor{forestgreen}{$+3.7$}& \textcolor{forestgreen}{$+3.4$}\\
        %%%%%%
        \multirow{2}{*}{Default Tip} & $60.0 \pm 0.2$ & $62.1 \pm 0.2$ & $63.8 \pm 0.2$ & $68.6 \pm 0.3$ & $69.5 \pm 0.2$\\
        & \textcolor{forestgreen}{$+5.4$} & \textcolor{forestgreen}{$+6.2$} & \textcolor{forestgreen}{$+5.4$} & \textcolor{forestgreen}{$+4.8$}& \textcolor{forestgreen}{$+4.5$}\\
        \multirow{2}{*}{Softmax NN} & $64.5 \pm 0.2$ & $66.8 \pm 0.1$ & $68.2 \pm 0.2$ & $72.0 \pm 0.2$ & $73.1 \pm 0.1$\\
        & \textcolor{forestgreen}{$+2.7$} & \textcolor{forestgreen}{$+2.3$} & \textcolor{forestgreen}{$+2.8$} & \textcolor{forestgreen}{$+2.8$} & \textcolor{forestgreen}{$+2.3$}\\
    \end{tabular}}
    \vspace{-8pt}
    \caption{\textbf{Context-aware pretraining for CLIP.} Benefits of contextualized pretraining also directly transfer to pretraining using the CLIP (softmax-based) objective~\cite{radford2021clip}, \clipours,  here shown for $32$-shot classification for three metric-based classifiers. } % \vspace{-1em}
    \label{tab:clip_arch_comp}
    \vspace{-8pt}
\end{table}

\vspace{0.5em} \noindent \textbf{Robustness to choice of pretraining objective.} While we conduct most of our experiments using the more scalable SigLIP~\cite{zhai2023siglip} objective, \cref{tab:clip_arch_comp} shows that all the performance gains hold when switching the underlying pretraining loss to CLIP \cite{radford2021clip}. For each architecture and example count, we directly transfer the same hyperparameters to context-aware pretraining using \clipours \ (\cref{eq:ctxsiglip}) with $\alpha=0.9$ for 1.5B and 6B training examples, and $\alpha=0.95$ for 15B runs. 
% Importantly, again, independently learned temperatures $\tau_i = \exp(\tau'_i)$ are utilized. 
Similarly to the \sigours \ results, we see significant improvements when varying backbones, training duration and adaptation methods --- e.g. \textcolor{forestgreen}{$+6.2\%$} for a Tip-Adapter applied on a ViT-S/16 (6B) model or \textcolor{forestgreen}{$+2.8\%$} for a softmax-voted nearest neighbor classifier on a ViT-S/16 (15B) model --- while maintaining the base zero-shot transfer performance. 
% (\textcolor{forestgreen}{$+0.0\%$} and \textcolor{firebrickred}{$-0.1$}, respectively). 
This highlights the \textit{generality of context-aware pretraining} for different types of vision-language contrastive pretraining.\\

\begin{figure}[t!]
    \centering
    \includegraphics[width=1\linewidth]{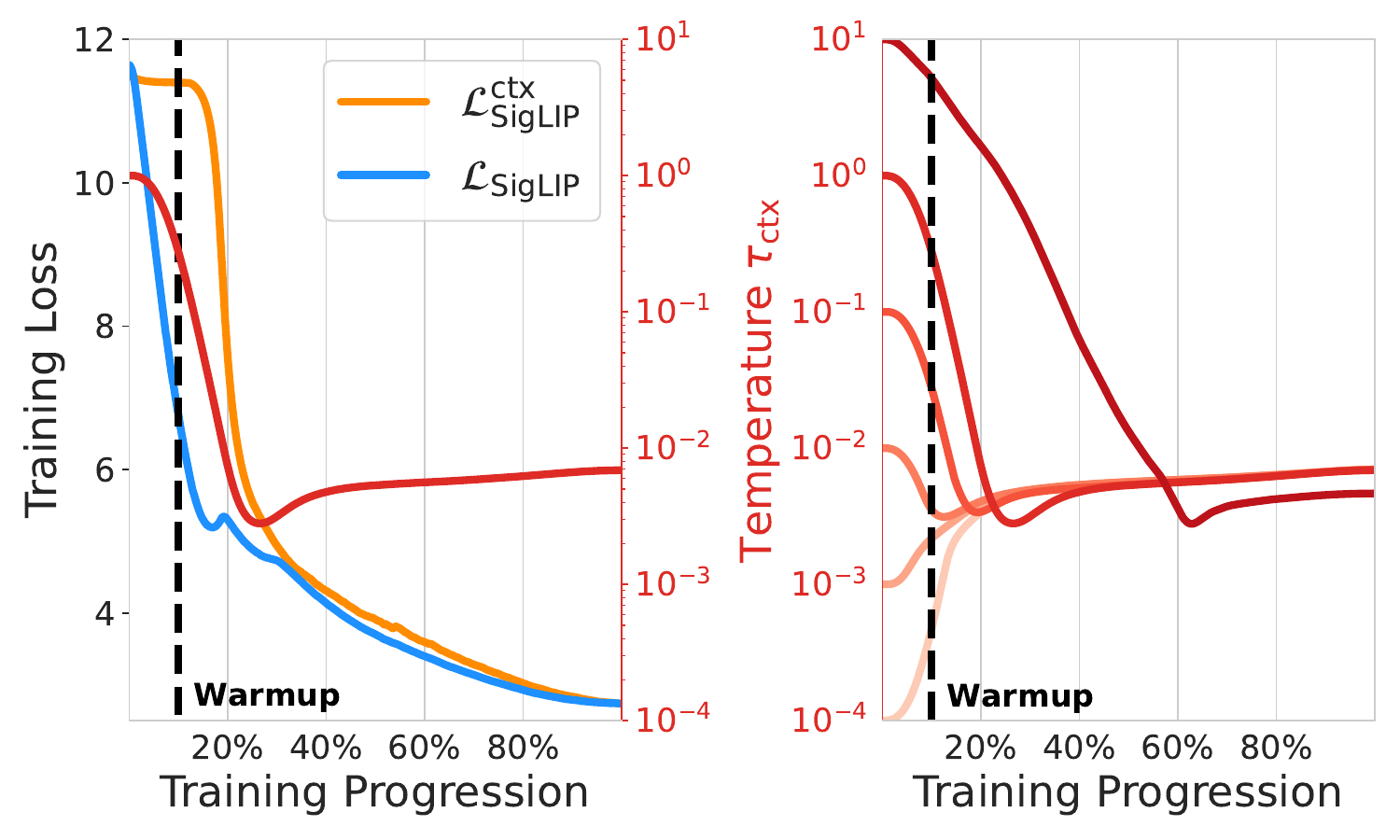}
    \vspace{-20pt}
    \caption{\textbf{Training Dynamics.} \textit{(Left)} Relation between the \textcolor{darkerblue}{base SigLIP training objective} $\mathcal{L}_\text{SigLIP}$ and its \textcolor{darkorange}{contextualized counterpart} $\mathcal{L}_\text{SigLIP}^\text{ctx}$ (here referring to the second term in \cref{eq:ctxsiglip}) for a ViT-B/16 model trained on 6B WebLI examples using \sigours. We jointly visualize the learned \textcolor{firebrickred}{contextualization temperature $\tau_\text{ctx}$} (note the log-scale). \textbf{(Right)} Progression of \textcolor{firebrickred}{$\tau_\text{ctx}$} for different initializations. We find that context-usage is emergent during training and dependent on a suitable learned \textcolor{firebrickred}{context temperature $\tau_\text{ctx}$}, which fortunately is very robust across initializations.}
    \label{fig:loss_relations}
    \vspace{-8pt}
\end{figure}
\begin{table*}[t!]
    \centering
    \small

\begin{subtable}[h!]{0.3\linewidth}
    \centering
    \resizebox{1\linewidth}{!}{
    \begin{tabular}{lcc}
        Method & Zero-Shot & 16-Shot \\
        \midrule
        \rowcolor{lightblue}
        \texttt{SigLIP} & $51.0$ & $60.1 \pm 0.4$ \\
        \rowcolor{lightorange}
        \texttt{Sig\textbf{LIxP}} & $50.5$ & $64.1 \pm 0.5$ \\
        No masking & $50.9$ & $60.5 \pm 0.4$ \\        
        ~ \\
    \end{tabular}}
\caption{\textbf{Self-Attention Masking}\vspace{1em}}
\label{tab:ablations_method_1_1}
\end{subtable}
~
\begin{subtable}[h!]{0.3\linewidth}
    \centering
    \resizebox{1\linewidth}{!}{
    \begin{tabular}{lcc}
        Method & Zero-Shot & 16-Shot \\
        \midrule
        $\alpha=0.95$ & $50.8$ & $62.5 \pm 0.3$ \\
        \rowcolor{lightorange}
        $\alpha=0.9$ & $50.5$ & $64.1 \pm 0.5$ \\
        $\alpha=0.8$ & $50.0$ & $63.8 \pm 0.3$ \\
        $\alpha=0.6$ & $48.7$ & $61.5 \pm 0.4$ \\
    \end{tabular}}
\caption{\textbf{Relative weighting}\vspace{1em}}
\label{tab:ablations_method_1_2}
\end{subtable}
~
\begin{subtable}[h!]{0.3\linewidth}
    \centering
    \resizebox{1\linewidth}{!}{
    \begin{tabular}{lcc}
        Method & Zero-Shot & 16-Shot \\
        \midrule
        \rowcolor{lightorange}
        Single-Stage & $50.5$ & $64.1 \pm 0.5$ \\
        Residual & $49.2$ & $59.2 \pm 0.4$ \\
        Multimodal & $47.9$ & $61.4 \pm 0.4$ \\
        Two-Stage & $50.5$ & $62.0 \pm 0.3$ \\
        % (d) Multimodal TS $m=2$ & $50.0$ & $63.8 \pm 0.3$ \\
    \end{tabular}}
\caption{\textbf{Contextualization Type}\vspace{1em}}
\label{tab:ablations_method_1_3}
\end{subtable}

% \multicolumn{3}{l}{\textbf{(Part 3) Contextualization Type} (using best $\alpha$)}\\
% (a) Residual connection & $49.2$ & $59.2\pm0.4$\\
% (b) Image Keys, Text Values & $47.9$ & $61.4 \pm 0.4$ \\
% (c) Two-Stage $m=2$ & $50.5$ & $62.0\pm0.3$\\
% (d) Hummingbird & $48.5$ & $61.5 \pm 0.4$ \\
        
\begin{subtable}[h!]{\columnwidth}
    \centering
    \resizebox{0.75\linewidth}{!}{
    \begin{tabular}{lcc}
        Method & Zero-Shot & 16-Shot \\
        \midrule
        \rowcolor{lightorange}
        $\tau_1, \tau_2, \tau_\text{ctx}$ learnable & $50.5$ & $64.1 \pm 0.5$ \\        
        $\tau_1 = \tau_2, \tau_\text{ctx}$ & $47.8$ & $61.8 \pm 0.4$ \\
        $\tau_1, \tau_2 = \tau_\text{ctx}$ & $50.4$ & $60.2 \pm 0.6$ \\
        $\tau_\text{ctx}$ frozen & $47.1$ & $59.4 \pm 0.5$\\
    \end{tabular}}
\caption{\textbf{Uncoupled, learnable temperatures}}
\label{tab:ablations_method_1_4}
\end{subtable}
~
\begin{subtable}[h!]{\columnwidth}
    \centering
    \resizebox{0.75\linewidth}{!}{
    \begin{tabular}{lcc}
        Method & Zero-Shot & 16-Shot \\
        \midrule
        \rowcolor{lightorange}
        Full back-propagation & $50.5$ & $64.1 \pm 0.5$ \\       
        Stop Gradient: $\{K\}$ & $49.6$ & $62.7 \pm 0.4$ \\
        Stop Gradient: $\{V\}$ & $43.8$ & $58.7 \pm 0.2$ \\
        Stop Gradient: $\{K, V\}$ & $44.4$ & $56.5 \pm 0.4$ \\
    \end{tabular}}
\caption{\textbf{Back-propagation through memory buffer}}
\label{tab:ablations_method_1_5}
\end{subtable}
\vspace{-8pt}
\caption{\textbf{Context-Aware Pretraining Ablations.} We ablate the context-aware training objective in \cref{eq:ctxsiglip} and \cref{eq:final_ctx} on a ViT-S/16 (1.5B), evaluated with prototypical classifiers on a subset of our evaluation benchmarks. We report average zero- and 16-shot performances.%Results clearly motivate the need for masking $\mathbf{M}$ in \cref{eq:final_ctx} \textbf{(1)}, robustness towards relative weighting $\alpha\in[0.8, 0.95]$ \textbf{(3)}, our base contextualization setup in \cref{eq:final_ctx} versus more complex, multimodal instantiations \textbf{(3)}, temperature-based separation of losses in \cref{eq:ctxsiglip} and \cref{eq:final_ctx} \textbf{(4)}, and the importance of full backpropagation in \cref{eq:final_ctx} \textbf{(5)}.
\vspace{-1em}
}
\label{tab:ablations_method_1}
    
\end{table*}

\noindent\textbf{Contextualized Pretraining Dynamics.} 
To understand contextualized pretraining dynamics, we compare the dynamics of the \textcolor{darkerblue}{base SigLIP training loss $\mathcal{L}_\text{SigLIP}$} and the \textcolor{darkorange}{contextualized SigLIP counterpart $\mathcal{L}_\text{SigLIP}^\text{ctx}$} (\cref{eq:ctxsiglip}, first and second terms). \Cref{fig:loss_relations} (left) visualizes this against the changes in \textcolor{firebrickred}{contextualization temperature $\tau_\text{ctx}$}. During early stages of training (around the warmup period), no context use is learned. Only once the \textcolor{darkerblue}{base objective} reaches a certain threshold with sufficient representation quality does the model start to leverage available context following \cref{eq:final_ctx}. 
This coincides with \textcolor{firebrickred}{$\tau_\text{ctx}$} reaching a value sufficiently low to encourage strong separation of buffer entries. During this joint inflection point, we often find an increase in the \textcolor{darkerblue}{base siglip loss}, followed by a slight increase in \textcolor{firebrickred}{$\tau_\text{ctx}$} for less aggressive context selection to better align the base representation pretraining and the contextualization paradigm.
We find that ensuring \textcolor{firebrickred}{$\tau_\text{ctx}$} to be exponentially learnable following the formulation using in \citet{zhai2023siglip} is crucial to effectively trade-off retention  of base zero-shot transfer capabilities and the improvement in few-shot adaptability.
This exponential treatment also introduces robustness towards \textcolor{firebrickred}{$\tau_\text{ctx}$} starting values (\cref{fig:loss_relations}, right); with temperatures converging to similar inflection points irrespective of initializations across orders of magnitude ($10^{-4}$ to $1$). % Only for much higher temperature values are training dynamics more severely impact; delaying the inflection point and the emergence of context use.

\begin{table*}[t!]
\centering
\begin{subtable}[h!]{0.32\linewidth}
    \centering
    \resizebox{1\linewidth}{!}{
    \begin{tabular}{lcc}
        Method & Zero-Shot & 16-Shot \\
        \midrule
        \rowcolor{lightblue}
        \texttt{SigLIP} & $51.0$ & $60.1 \pm 0.4$ \\
        \rowcolor{lightorange}
        \texttt{Sig\textbf{LIxP}} & $50.5$ & $64.1 \pm 0.5$ \\
        Separate Batch & $48.8$ & $63.2 \pm 0.3$ \\        
        ~ \\
    \end{tabular}}
\caption{\textbf{Context Batch Separation}\vspace{1em}}
\label{tab:ablations_method_2_1}
\end{subtable}
~
\begin{subtable}[h!]{0.32\linewidth}
    \centering
    \resizebox{1\linewidth}{!}{
    \begin{tabular}{lcc}
        Method & Zero-Shot & 16-Shot \\
        \midrule
        \rowcolor{lightorange}
        None & $50.5$ & $64.1 \pm 0.5$ \\
        linear & $50.2$ & $62.8 \pm 0.2$\\
        2-layer MLP & $49.8$ & $61.1 \pm 0.2$\\
        % V MLP (2-layer, large) & $50.0$ & $61.9 \pm 0.3$\\
        3-layer MLP & $49.1$ & $60.8 \pm 0.3$\\
    \end{tabular}}
\caption{\textbf{Value Heads for $\mathcal{M}_V$}\vspace{1em}}
\label{tab:ablations_method_2_2}
\end{subtable}
~
\begin{subtable}[h!]{0.32\linewidth}
    \centering
    \resizebox{1\linewidth}{!}{
    \begin{tabular}{lcc}
        Method & Zero-Shot & 16-Shot \\
        \midrule
        \rowcolor{lightblue}
        \texttt{SigLIP} & $51.0$ & $60.1 \pm 0.4$ \\
        \rowcolor{lightorange}
        \texttt{Sig\textbf{LIxP}} & $50.5$ & $64.1 \pm 0.5$ \\
        No Normalization & $48.9$ & $61.7 \pm 0.3$ \\    
        ~ \\
    \end{tabular}}
\caption{\textbf{QK Normalization}\vspace{1em}} % Removing 
\label{tab:ablations_method_2_3}
\end{subtable}
~        
\begin{subtable}[h!]{0.32\linewidth}
    \centering
    \resizebox{1\linewidth}{!}{
    \begin{tabular}{lcc}
        Method & Zero-Shot & 16-Shot \\
        \midrule
        \rowcolor{lightorange}
        None & $50.5$ & $64.1 \pm 0.5$ \\
        LayerNorm $\{K\}$ & $46.8$ & $57.1 \pm 0.4$\\
        LayerNorm $\{V\}$ & $48.5$ & $62.3 \pm 0.5$\\
        LayerNorm $\{K, V\}$ & $48.3$ & $58.2 \pm 0.3$\\
    \end{tabular}}
\caption{\textbf{Layer Normalization on $\mathcal{M}$}}
\label{tab:ablations_method_2_4}
\end{subtable}
~
\begin{subtable}[h!]{0.32\linewidth}
    \centering
    \resizebox{1\linewidth}{!}{
    \begin{tabular}{lcc}
        Method & Zero-Shot & 16-Shot \\
        \midrule
        \rowcolor{lightorange}
        None & $50.5$ & $64.1 \pm 0.5$ \\       
        + Stale (32K) & $45.9$ & $59.7 \pm 0.3$\\
        + Stale (128K) & $46.9$ & $60.5 \pm 0.5$\\
        + Stale (512K) & $47.2$ & $60.7 \pm 0.4$\\
    \end{tabular}}
\caption{\textbf{Inclusion of Stale Buffer}}
\label{tab:ablations_method_2_5}
\end{subtable}
~
\begin{subtable}[h!]{0.32\linewidth}
    \centering
    \resizebox{1\linewidth}{!}{
    \begin{tabular}{lcc}
        Method & Zero-Shot & 16-Shot \\
        \midrule
        \rowcolor{lightorange}
        Full (32k) & $50.5$ & $64.1 \pm 0.5$ \\       
        Subset (1k) & $50.7$ & $59.9 \pm 0.5$ \\
        Subset (2k) & $50.5$ & $62.9 \pm 0.4$ \\
        Subset (8k) & $50.3$ & $63.9 \pm 0.3$ \\
    \end{tabular}}
\caption{\textbf{Reduced Active Buffer Size}}
\label{tab:ablations_method_2_6}
\end{subtable}
\vspace{-10pt}
\caption{\textbf{Context Buffer Ablations.} Following \cref{tab:ablations_method_1}, we study context buffer design $\mathcal{M}_K$ and $\mathcal{M}_V$. Our results clearly motivate our most scalable, and yet best performing variant in \cref{eq:final_ctx} which directly reuses batch-level image representations $\phi_I(\mathcal{B}_I)$.}
\label{tab:ablations_method_2}
\vspace{-8pt}
\end{table*}

\subsection{Ablations}\label{subsec:ablations}
The context-aware pretraining framework introduced in \cref{subsec:contextualized_pretrain} is the best performing, most scalable instance of a larger contextualization framework which we explored. In this section, we characterize the design space for our context-aware pretraining objective in \cref{tab:ablations_method_1}, and for particular buffer design choices $\mathcal{M}_K$ and $\mathcal{M}_V$ in \cref{tab:ablations_method_2}. For more details, see supplementary.\\

\noindent \textbf{Mitigating shortcut solutions.} We first show the importance of correctly masking out self-attention entries in the cross-attention formalism in \cref{eq:final_ctx} via $\mathbf{M}$ (\cref{tab:ablations_method_1_1}). We can see that without masking out self-attention in \cref{eq:final_ctx}, no context re-use emerges with results matching the baseline SigLIP (first row), as the model simply learns to attend to itself (having the highest similarity by default), consequently having no incentive to leverage other context.\\

\noindent \textbf{Loss balancing.} 
% With \cref{fig:loss_relations} showing high robustness to contextualization temperature initializations $\tau_\text{ctx}$ and SigLIP temperatures directly adapted from \cite{zhai2023siglip}, the main hyperparameter to tune 
We ablate the relative weighting $\alpha$ between base image-text contrastive objective and its contextualized counterpart (\cref{eq:ctxsiglip}) in \cref{tab:ablations_method_1_2}, observing significant gains in few-shot transfer and retention of base zero-shot performance for $\alpha\in[0.95, 0.8]$.
For larger $\alpha$, training regresses back to the base image-text contrastive objective, while smaller $\alpha$ skew too heavily towards context usage at increased cost to zero-shot transfer.\\

% We do note that relative to the respective zero-shot performance, larger contextualization weights incur highest context usage relatively speaking (e.g. for $\alpha=0.8$, zero-shot performance goes down from $51.0\%$ to $45.2\%$, while $16$-shot performance only goes down from $60.1\%$ to $59.4\%$).\\

\noindent \textbf{Contextualization objective.} 
% By default, we conduct a single contextualization step with $\mathcal{M}_K \!=\! \mathcal{M}_V \!=\! \mathcal{B}_I$. In this part, we investigate multiple contextualization steps, as well as modifications to the structure of \cref{eq:final_ctx}. 
In \cref{tab:ablations_method_1_3}, we investigate several modifications to the contextualization objective.
\textbf{(a)} Instead of separating training into base and context loss (c.f.~\cref{eq:ctxsiglip}), we instead utilize a residual connection to \cref{eq:context} and train on a single objective $\mathcal{L}_\text{SigLIP}\left(\alpha{}x_i+(1-\alpha)x^\text{ctx}_i, t_i, \tau_1\right)$.
\textbf{(b)} Multimodal contextualization by setting $\mathcal{M}_V$ to use textual representations $\mathbf{T}$.
\textbf{(c)} Conducting two successive contextualization steps via
\begin{equation}\label{eq:successive_context}
    x^\text{ctx}_{i,m} \!=\! \psi_\text{ctx}\left(f_\text{ctx}(x^\text{ctx}_{i,m-1}, \mathcal{M}^{m-1}_K, \mathcal{M}^{m-1}_V\right))    
\end{equation}
with $m\in\{1, ..., M\}, M=2,$ and $f_\text{ctx}$ our contextualization objective from \cref{eq:final_ctx}. $\psi_\text{ctx}$ defines an optimal learnable map, which we set to a linear one for our two-stage contextualization. Note that for $M=1$ and $\psi_\text{ctx}$ the identity function we recover our base objective in \cref{eq:final_ctx}.
%
% \textbf{(d)} A combination of \textbf{(b)} and \textbf{(c)} resulting in two-stage, multimodal contextualization,
% Hummingbird-style contextualization~\cite{balazevic2023hummingbird} 
% with $m=2$, the first-stage contextualization utilizing remapped image embeddings $\mathcal{M}^{m=1}_V=\psi_\text{2-Layer MLP}(\mathbf{X})$ and the second stage incorporating textual context $\mathcal{M}^{m=2}_V=\mathbf{T}$. Again, $\psi_\text{ctx}$ denotes an intermediate linear map.\\
% 
% \noindent
First, we find that for residual connection in \textbf{(a)}, the optimal $\alpha$ was $\alpha=1$, i.e. effectively retaining the base contrastive training objective. For slightly smaller values, i.e capping $\alpha$ at $0.9$, both zero-shot transfer and few-shot adaptability (\textcolor{darkerblue}{$51.0\%$}$\rightarrow{}49.2\%$ and \textcolor{darkerblue}{$60.1\%$}$\rightarrow{}59.2\%$) are negatively impacted, as it effectively mitigates effectiveness of the base SigLIP objective. This showcases the \textit{explicit need for separation of base and contextualization losses} alongside distinctly learnable temperatures. Utilizing multimodal context \textbf{(b)} reduces transfer performance, albeit maintaining slight gains in few-shot capabilities, and only works at all if masking $\mathbf{M}$ in \cref{eq:final_ctx} is set, as otherwise a simple weight-copying shortcut is found. Finally, we find that a two-stage contextualization \textbf{(c)} reduces few-shot gains compared to \textit{single-stage contextualization}.

\vspace{0.5em} \noindent \textbf{Individual, learnable temperatures.} Having three distinct temperatures ($\tau_1$, $\tau_2$, $\tau_\text{ctx}$) is crucial for optimal performance as shown in \cref{tab:ablations_method_1_4}. We ablate \textbf{(a)} sharing the temperature across all constituents of \cref{eq:ctxsiglip}, \textbf{(b)} sharing the \sigours \ and contextualization temperature, and \textbf{(c)}
% utilizing a single learnable temperature; and \textbf{(d)}
freezing $\tau_\text{ctx}$ to either $\{0.01, 0.1, 1\}$ and reporting the best results. We find that allowing $\tau_\text{ctx}$ to adapt to be crucial for few-shot gains and restricting $\tau_1 = \tau_2$ to negatively impact the zero-shot transfer ability. Consequently, our results strongly advocate for a \textit{full set of independently learnable} temperatures.

\vspace{0.5em} \noindent \textbf{Key and value backpropagation.} \cref{eq:final_ctx} allows the model backpropagate through key and value entries $\mathcal{M}_K$ and $\mathcal{M}_V$. To understand which component is most crucial for peak performance, we ablate the gradient flow through $\mathcal{M}_K$ and $\mathcal{M}_V$ in \cref{tab:ablations_method_1_5}. Interestingly, freezing $\mathcal{M}_V$ and only optimizing for the retrieval component (i.e. entries within $\sigma(\cdot)$ in \cref{eq:final_ctx}, directly mimicking retrieval operations at test-time) is highly detrimental (16-shot score \textcolor{darkorange}{$64.1\%$} $\rightarrow$ $58.7\%$), whereas freezing $\mathcal{M}_K$ has a much smaller negative impact ($62.7\%$ versus \textcolor{darkerblue}{$60.1\%$}).
For contextualization to function for pretraining, allowing the model to optimize for both key, but especially value entries is thus important to reach optimal context usage.\\
% Consequently, one could argue that allowing the model to optimize how it should benefit from context retrieval is more important that optimization of the actual retrieval process; though in practice, both are crucial to reach the final peak performance.\\

\noindent
\textbf{General Buffer Design.} To understand the right way to structurally design the contextualization buffer (see \Cref{tab:ablations_method_2}), we investigate using examples from a separate batch $\mathcal{B}_I$ to populate $\mathcal{M}$ (\cref{tab:ablations_method_2_1}), learnable value heads over $\mathcal{M}_V$ (\cref{tab:ablations_method_2_2}), normalization of query-key representations in \cref{eq:final_ctx} (\cref{tab:ablations_method_2_3}), layer normalization of buffer entries (\cref{tab:ablations_method_2_4}),  the inclusion of stale memories from previous iterations (\cref{tab:ablations_method_2_5}), and finally the reduction of utilized $\mathbf{X}_I$ in \cref{eq:final_ctx} (\cref{tab:ablations_method_2_6}).
% \\
% 
% \noindent
\Cref{tab:ablations_method_2_1} shows that populating $\mathcal{M}_K$ and $\mathcal{M}_V$ with a separately embedded batch results in a small decrease in both zero-shot and few-shot performance. From \Cref{tab:ablations_method_2_2} and \Cref{tab:ablations_method_2_4}, we conclude that no value heads or layer normalization are needed - most closely aligning with the direct re-use of learned embeddings downstream. Similarly, \cref{tab:ablations_method_2_3} shows that normalized embeddings for QK weighting in \cref{eq:final_ctx} are essential. 
% (although even without it, we gains in few-shot performance are possible; e.g. going from \textcolor{darkerblue}{$60.1\%$} to $61.7\%$). 
Extending $\mathcal{M}$ with stale representations from previous iterations in \cref{tab:ablations_method_2_5} offers no benefits, though on the other hand, maintaining a large enough active buffer $|\mathcal{M}|=|\mathcal{B}|$ in \textbf{(f)} is crucial to reliably benefit from contextualized training.
% , where having the full batch of 32k examples outperforms the alternatives. 
% Indeed, setting the active buffer to small impacts training stability and mitigates gains in few-shot adaptation performance.

\section{Conclusion}\label{sec:conclusion}
We introduced a context-aware pretraining objective for large-scale vision-language representation learning that facilitates few- and many-shot visual context use in a training-free, metric-based manner at test time. Importantly, we demonstrate the possibility of significantly boosting visual adaptation performance with no compromises in underlying zero-shot transfer capabilities. Extensive evaluations on 21 diverse visual adaptation tasks show up to \textit{four-fold} gains in test-time sample efficiency and improvements in average few-shot performance by often over $5\%$, closing the gap to more complex optimization-based strategies. These gains hold across model and data scales, highlighting the potential of simple and scalable pretraining strategies to augment test-time adaptation beyond simple zero-shot transfer.

\section*{Acknowledgements}\label{sec:ack}
The authors would like to thank Nikhil Parthasarathy and Relja Arandelović for helpful feedback. KR thanks the International Max Planck Research School for Intelligent Systems (IMPRS-IS), the European Laboratory for Learning and Intelligent Systems (ELLIS) PhD program and the ELISE Mobility Grant for support.
ZA acknowledges the support from the German Research Foundation (DFG): SFB 1233, Robust Vision: Inference Principles and Neural Mechanisms, project number: 276693517 and ERC Grant DEXIM, project number: 853489. All authors thank Google DeepMind for providing the resources and high-quality environment for this research.

%%%%%%%% Supplementary & Bibliography
% \clearpage
{
    \small
    \bibliographystyle{ieeenat_fullname}
    \bibliography{main}
}

\appendix

\clearpage
\setcounter{page}{1}
\maketitlesupplementary

% Things to include:
% \begin{itemize}
%     % \item Discussion of limitations and societal impact.
%     \item Include dataset table.
%     \item Additional plots for other architectures.
%     \item WIP: LiT Decoder experiments.
%     \item WIP: Additional generalization experiments.
%     \item WIP: Latent space exploration results.
% \end{itemize}

% \section{Limitations and Broader Impact}
% \label{suppsec:limitations}

\section{Experimental Details}
\label{sec:supp_experimental_details}
In this section, we provide the complete experimental details for our primary training runs.
In particular, we study various data scales on WebLI~\cite{chen2023pali}, ranging from least 1.5 billion up to 15 billion training examples. Our batchsize is set to 32768 by default following optimal suggestions in \citet{zhai2023siglip}. Images for training are resized to $256\times 256$. We use Adam-W optimizer~\cite{loshchilov2018adamw} with learning rate of $10^{-3}$, weight decay of $10^{-4}$, gradient clipping to norm $1$, and $\beta_2 = 0.95$ following recommendations in \cite{zhai2023siglip,evans2024datacurationjointexample}.
The full pipeline is implement in \texttt{jax}~\cite{jax2018github}.
Our vision encoder is parameterized as a vision transformer~\cite{dosovitskiy2021vit}. The corresponding text-encoder (a standard transformer ~\cite{vaswani2017transformer}) tokenizes input text using the sentencepiece tokenizer~\cite{kudo2018sentencepiece} pretrained on the English C4 dataset~\cite{raffel2020c4}. If not noted otherwise, \textbf{LIxP}-training utilizes $\alpha=0.9$, $\tau_\text{ctx} = 1$, $\tau_1 = 10$ following \cite{zhai2023siglip}, and $\tau_2 = \tau_1$. While more detailed hyperparameter grid searches would likely provide even better results, we opt for a simple and transferable parameter grid for easiest reuse and replication.
To evaluate both the zero-shot transfer capabilities as well as the few-shot adaptation performance, we measure performance on 21 diverse datasets commonly used for few-shot and domain adaptation works: CUB200-2011~\cite{cub2002011}, Stanford Cars~\cite{cars196}, Cassave~\cite{cassava}, CIFAR100~\cite{cifar100}, Colorectal Histology~\cite{colhist}, DomainNet-$\{$ClipArt, Infograph, Quickdraw, Sketch$\}$~\cite{domainnet}, DTD~\cite{dtd}, EuroSAT~\cite{eurosat}, Food101~\cite{food101}, ImageNet2012~\cite{imagenet2012}, ImageNet-Sketch~\cite{imagenetsketch}, Oxford IIIT Pets~\cite{pets}, Places365~\cite{places365}, Plant-Village~\cite{plantvillage}, RESISC45~\cite{resisc45}, Stanford Dogs~\cite{dogs}, SUN397~\cite{sun397} and UC Merced~\cite{ucmerced}.
Datasets are selected to allow for shot counts of at least up to 28-32, and were queried through the \texttt{tensorflow datasets} interface, see \url{tensorflow.org/datasets/catalog}. For datasets where only a single split was available (such as only train or test), we create a support/test split to allow for sufficient adaptation examples, but ensuring that the number of classes are maintained. The exact splits are provided in \cref{supp_tab:data_splits}. Ablation runs are evaluated on a subset (eleven, $\approx$half) of our evaluation benchmarks, and cover: CUB200-2011, Stanford Cars, Colorectal Histology, DTD, EuroSAT, Food101, ImageNet2012, ImageNet-Sketch, Oxford IIIT Pets, Places365 and UC Merced --- reporting average 16-shot performance.

\section{Nearest-neighbor voting classifiers}
\label{sec:supp_nn_voting}
As described in the main part of this paper, we study multiple different instantiations of nearest-neighbor classifiers based on varying neighbor sample weights $w_i$. These are:

\paragraph{Plurality-Voted Nearest-Neighbor Classifier, e.g. \cite{nakata2022knn}.} 
% Given $\mathbf{X}_\text{spt}$ and the corresponding labels $L_\text{spt}$, 
We compute $k$ nearest neighbors $\mathbf{X}_\text{spt}^k$ to $x_\text{test}$ and the label for $x_\text{test}$ is computed as the majority label from the corresponding labels $L_\text{spt}^k$. For all our experiments with plurality voting, we fix $k=32$, but capped to the maximum number of shots for a given few-shot classification task.

\paragraph{Softmax-Voted Nearest-Neighbor Classifier, e.g. \cite{caron2021dino,geirhos2024flexibleperceptionvisualmemory}} For each of the $k$ nearest neighbors $\mathbf{X}^{k}_\text{spt}$ with respect to $x_\text{test}$, we assign a softmax sample weight for the $i$-th neighbor in $\mathbf{X}^k_\text{spt}$ (with temperature $\tau_s$) as
\begin{equation}\label{eq:softmaxvoting}
    w_i = \frac{\exp(x_q\mathbf{X}^k_{\text{spt},i}/\tau_s)}{\sum_{j=1}^{k}\exp(x_q\mathbf{X}^k_{\text{spt},j}/\tau_s)}.
\end{equation}
We follow existing literature~\cite{wu2018instance,caron2021dino,oquab2024dinov2,geirhos2024flexibleperceptionvisualmemory} and keep $\tau_s=0.07$, while setting $k=32$. The final output logits are then simply computed as the softmax-weighted aggregation of the one-hot labels $\mathbf{L}^k_\text{spt}$ of the neighbors.

\paragraph{Rank-Voted Nearest-Neighbor Classifier~\cite{geirhos2024flexibleperceptionvisualmemory}.} This nearest-neighbor classifier computes the weights of $k$-neighbors following a simple power-function
% \begin{equation}\label{eq:rankvoting}
    $w_i = 1 / (\gamma + \text{rank}_i)$
% \end{equation}
with offset $\gamma = 2.0$~\cite{geirhos2024flexibleperceptionvisualmemory}, and rank of support image index $i$ $\mathbf{X}^k_{\text{spt},i}$ within the $k$ neighborhood.

\section{Additional Buffer Studies}
\label{sec:supp_additional_results}
% \subsection{Shot-Scaling Plots for Other Architectures}
% \subsection{Shot-Scaling Plots for C\textbf{LIxP}-trained Models}
% \subsection{Augmented Input Buffer}
\begin{table}[t!]
    \centering
    \resizebox{\linewidth}{!}{
    \begin{tabular}{lcc}
        \textbf{Method} & \textbf{Avg. Zero-Shot} &\textbf{ Avg. 16-Shot} \\
        \midrule
        % \rowcolor{lightblue}
        % \texttt{SigLIP} ($\alpha=1$) & $51.0$ & $60.1 \pm 0.4$ \\
        \rowcolor{lightorange}
        No Augmentations & $50.5$ & $64.1 \pm 0.5$ \\
        % \multicolumn{3}{l}{\textbf{(g) Buffer Design}: Augmentation Entries}\\
        Augmented (Buffer only) & $48.3$& $60.0 \pm 0.3$\\
        Augmented (All) & $49.5$ & $63.8 \pm 0.3$\\
        Augmented (All) + InfoNCE & $51.0$ & $63.9 \pm 0.3$ \\
    \end{tabular}}
    \vspace{-5pt}
    \caption{\textbf{Additional Buffer Ablations:} Inclusion of augmented entries, with and without additional InfoNCE-style training augmenting the base image-text contrastive training.}
    \label{tab:supp_ablations_method}
\end{table}
We include an additional buffer design ablation, within which we study the option to populate the key and value contextualization buffer with augmented variants (``\textit{Augmented Entries}'') of the input batch $\mathcal{B}_I$ (and consequently removing the self-attention mask $\mathbf{M}$). In this scenario, we distinguish between only populating the buffer with augmented examples (``\textit{Buffer Only}''), as well as jointly training on them with and without the addition of a separate InfoNCE objective. Our results show that gains are only visible if augmented examples are treated as independent entries, effectively mimicking our main contextualization objective in \cref{eq:final_ctx}.

\begin{table*}[t!]
    \centering
    \resizebox{1\linewidth}{!}{
    \begin{tabular}{lll|ccc}
         \toprule
         \textbf{Dataset} & \textbf{Type} & \textbf{Support/Test Split} & \textbf{Support Examples} & \textbf{Test Set Size} & \textbf{\#Classes} \\
         \midrule
         CUB200-2011~\cite{cub2002011} & Finegrained, Birds & train, test & 5994 & 5794 & 200\\
         Stanford Cars~\cite{cars196} & Finegrained, Cars & train, test & 8144 & 8041 & 196\\
         Cassava~\cite{cassava} & Cassava Leafs & train, test & 5656 & 1885 & 5 \\
         CIFAR100~\cite{cifar100} & Visual Recognition & train, test & 50000 & 10000 & 100 \\
         Col. Histology~\cite{colhist} & Colorectal Cancer Histology & train[:2000], train[:2000] & 2000 & 3000 & 8\\
         DomainNet - ClipArt~\cite{domainnet} & Visual Recognition, ClipArt & train[:30K], test[:20K] & 30000 & 20000 & 345\\
         DomainNet - Infograph~\cite{domainnet} & Visual Recognition, Infographics & train[:30K], test[:20K] & 30000 & 20000 & 345\\
         DomainNet - Quickdraw~\cite{domainnet} & Visual Recognition, Quickdraws & train[:30K], test[:20K] & 30000 & 20000 & 345\\
         DomainNet - Sketch~\cite{domainnet} & Visual Recognition, Sketches & train[:30K], test[:20K] & 30000 & 20000 & 345\\
         DTD~\cite{dtd} & Textures & train, test & 1880 & 1880 & 47\\
         EuroSAT~\cite{eurosat} & Remote Sensing & train[:22K], train[22K:]& 22000 & 5000 & 10\\
         Food101~\cite{food101} & Finegrained, Food & train[:30K], validation & 30000 & 25250 & 101 \\
         ImageNet2012~\cite{imagenet2012} & Visual Recognition & train[:100K], validation & 100000 & 50000 & 1000 \\
         ImageNet-Sketch~\cite{imagenetsketch} & Visual Recognition, Sketch & test[:30K], test[35K:] & 30000 & 15889 & 1000 \\
         Oxford IIIT Pets~\cite{pets} & Finegrained, Pets & train, test & 3680 & 3669 & 37\\
         Places365 (small)~\cite{places365} & Finegrained, Places & train[:20K], validation[:15K] & 20000 & 15000 & 365\\
         Plant-Village~\cite{plantvillage} & Finegrained, Plant leaves & train[:30K], train[30K:] & 30000 & 24303 & 38 \\
         RESISC45~\cite{resisc45} & Remote Sensing & train[:20K], train[20K:] & 20000 & 11500 & 45\\
         Stanford Dogs~\cite{dogs} & Finegrained, Dogs & train, test & 12000 & 8580 & 120\\
         SUN397~\cite{sun397} & Scene Understanding & train[:30K], validation & 30000 & 10875 & 397 \\
         UC Merced~\cite{ucmerced} & Remote Sensing & train[:1K], train[1K:] & 1000 & 1100 & 21\\
         \bottomrule
    \end{tabular}}
    \caption{\textbf{Exact default support and test configurations} for all benchmark datasets studied. For most datasets with a clearly defined and available train and test split, we utilize these to define the pool of support examples to sample from for $K-shot$ few-shot studies, and the number of test examples evaluated on. For datasets (such as ``Col. Histology'' or ``Imagenet-Sketch'') where only one split was available through \texttt{tensorflow.datasets}, we split accordingly into support and test pool. For the remaining datasets (primarily DomainNet), we randomly subsample to maintain comparable support and test pools, though we note no relevant changes in relative performances across methods with either full or subsampled pools.}
    \label{supp_tab:data_splits}
\end{table*}

\end{document}